\documentclass{article}

\usepackage[preprint]{neurips_2026}

\usepackage[utf8]{inputenc}
\usepackage[T1]{fontenc}
\usepackage{hyperref}
\hypersetup{hidelinks} 
\usepackage{url}
\usepackage{booktabs}
\usepackage{amsfonts}
\usepackage{amsmath}
\usepackage{nicefrac}
\usepackage{microtype}
\usepackage[table]{xcolor}
\usepackage{multirow}
\usepackage{listings}
\usepackage{graphicx}
\usepackage{fvextra}
\usepackage{tabularx}

\newcommand{\promptinput}[2][]{%
  \VerbatimInput[
    fontsize=\footnotesize,
    breaklines=true,
    breakanywhere=true,
    breaksymbolright=\small$\hookrightarrow$,
    obeytabs=true,
    tabsize=2,
    frame=leftline,
    framerule=0.6pt,
    rulecolor=\color{gray},
    #1
  ]{#2}%
}
\usepackage{xspace}

\usepackage{mdframed}
\definecolor{rqboxbg}{gray}{0.93}
\definecolor{rqboxrule}{gray}{0.55}
\newif\ifrqfancy
\rqfancytrue 
\newmdenv[
  backgroundcolor=rqboxbg,
  linecolor=rqboxrule,
  linewidth=3pt,
  topline=false, bottomline=false, rightline=false, leftline=true,
  innerleftmargin=10pt, innerrightmargin=10pt,
  innertopmargin=7pt, innerbottommargin=7pt,
  skipabove=7pt, skipbelow=4pt,
]{rqbox}
\newenvironment{rqanswer}[1]{%
  \ifrqfancy\begin{rqbox}\fi\noindent\textbf{Answer to #1.}\ %
}{%
  \ifrqfancy\end{rqbox}\else\par\smallskip\fi%
}

\title{Auto-Configuring Scientific Simulators with Lightweight Coding-Agent Adapters}

\author{%
  Matthew Ho \quad Brian Liu \quad Jixuan Chen \quad Audrey Wang \quad Lianhui Qin \\[3pt]
  University of California, San Diego
}

\begin{document}

\maketitle

\begin{abstract}
Configuring an advanced scientific simulator, translating a modeling goal into a valid, runnable input deck, is a persistent bottleneck that costs domain scientists hours to days. Input decks are executable interfaces: simulator-specific vocabulary, cross-file references, schema constraints, and validation rules must align before a simulation can run. We show that this bottleneck can be substantially reduced with a lightweight adapter around an off-the-shelf coding agent, rather than a bespoke simulator agent. Coding agents already navigate files, edit code, run commands, and repair outputs; what they lack is the simulator’s executable contract, and rebuilding the agent loop risks discarding harness-calibrated tool-use and self-correction behavior. We introduce SIGA, a coding-agent adapter that supplies this contract through retrieval, procedural memory, agent-callable validation, and validation-gated termination while leaving the model and loop frozen. Because this contract is small and external, SIGA also supports adapter self-evolution: prior trajectories can rewrite the adapter contents without modifying the underlying agent. On GEOS, a multiphysics subsurface simulator, SIGA’s main gain is reliability: on harder held-out tasks it improves TreeSim from 0.720 to 0.789 and reduces across-run standard deviation by about 16× by preventing empty or invalid decks. In a human calibration, SIGA reaches in about five minutes the deck quality a domain expert reached in about three hours. Transfers to OpenFOAM and LAMMPS show the recipe is portable but interface-dependent: completion gates help when structural completeness is the bottleneck, while memory and retrieval help when value correctness is.

\end{abstract}

\section{Introduction}
\label{sec:intro}
\begin{figure}[t]
\centering
\includegraphics[width=\textwidth]{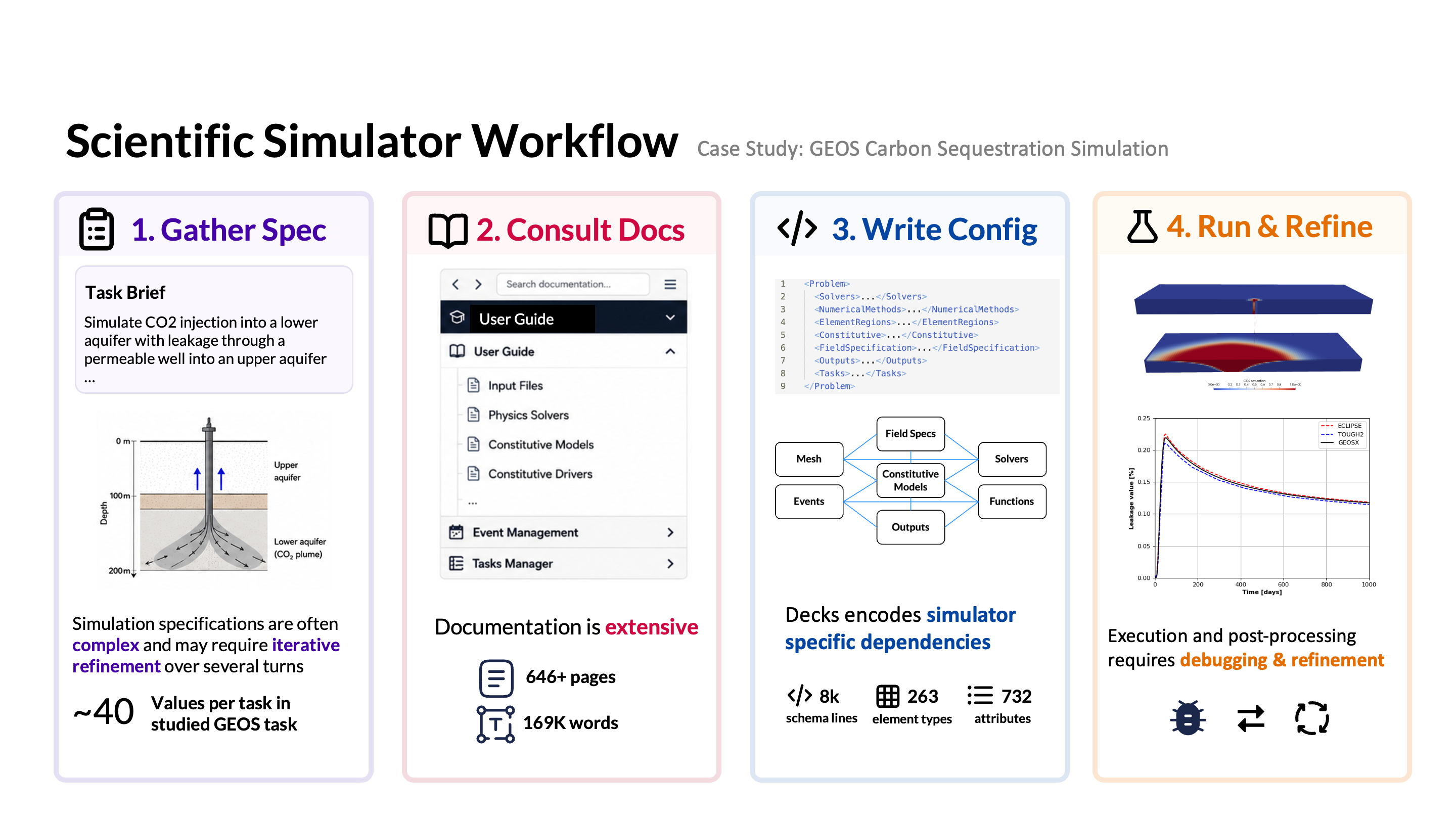}
\caption{Illustrative example of advanced tooling usage bottleneck for the geophysics domain. Here, the GEOS simulator's extensive documentation helps as a translation guide for its elaborate configuration that is a custom XML (domain specific language/DSL) to produce results such as simulating carbon sequestration in deep saline formations (top right) or reservoir flow in heterogeneous hydrocarbon (bottom right).}
\label{fig:1}
\end{figure}

Scientific simulation is one of the central computational workflows in modern science. Researchers use simulators to study physical, chemical, biological, and engineering processes that would be expensive, slow, dangerous, or impossible to observe directly in the laboratory. In geophysics, simulators such as the GEOS \footnote{\url{https://github.com/GEOS-DEV/GEOS}}~\citep{geos2024} model CO$_2$ sequestration, reservoir flow, hydraulic fracture, wellbore behavior, and induced seismicity; in molecular science, molecular-dynamics engines such as LAMMPS \citep{holbrook2026lammps} encode atomistic systems, force fields, and time-integration protocols. Across these domains, the scientific goal may be expressed in natural language, but the executable simulation must be written as a simulator specific input deck.

This translation from scientific intent to runnable simulator configuration is a persistent bottleneck in simulation-driven research. Advanced simulators are not simple tools with a few stable arguments. They expose large domain-specific language (DSL) whose tokens name solver classes, material models, mesh regions, boundary conditions, event schedules, numerical schemes, and output requests. A valid input deck must satisfy syntax constraints, schema constraints, and physical or domain-conventional constraints, while maintaining consistent names and references across sections and often across files. Scientists new to a simulator can therefore spend hours or days reading documentation, searching examples, testing numerical settings, and debugging invalid configurations before obtaining a runnable deck (Fig.~\ref{fig:1}).

We view simulator configuration as a valuable first target for AI for science because it is important, bounded, and verifiable. It is important because reducing setup time directly accelerates scientific iteration. It is bounded because the agent is not asked to invent new physics or interpret final scientific conclusions; it is asked to operate an existing scientific tool according to a specification. It is verifiable because the output is an artifact that can be parsed, validated, compared to reference decks, and eventually executed. If agents cannot reliably translate a specification into the input language of an existing simulator, they are unlikely to be dependable in broader scientific workflows that also require experimental design, hypothesis revision, and domain reasoning.

Recent scientific-agent systems have explored chemistry~\citep{bran2024chemcrow,boiko2023coscientist}, molecular dynamics~\citep{shi2026mdagent2,zhao2026polyjarvis,guilbert2025dynamate}, computational fluid dynamics~\citep{chen2024metaopenfoam,yue2025foamagent,pandey2025openfoamgpt}, finite-element and multiphysics modeling~\citep{zhan2025mooseagent,mcpsim2026,ni2024mechagents}, reservoir simulation~\citep{moyner2026jutulgpt}, and other simulator-centered workflows. Many of these systems build simulator specific agent loops from scratch, including custom planning, tool calling, retry logic, and termination handling. We study a different design point: adapt an existing coding agent rather than rebuild the agent. Modern coding agents already provide much of the generic machinery needed for simulator setup. They navigate repositories, inspect files, edit code, run commands, read errors, and repair outputs. What they lack is the simulator executable contract: the vocabulary, structural rules, validation procedures, and completion conditions that define a correct configuration. Preserving the native coding agent harness also preserves the tool use and self correction behavior that frontier models learn inside that harness (e.g.,~\citealp{cursor2026composer2}).

We introduce the Simulator Interface Grounding Adapter (SIGA), a lightweight coding agent adapter for automatic scientific simulator configuration. SIGA leaves the base model and agent loop frozen, and supplies simulator grounding only through the harness interfaces where setup failures occur. Retrieval gives semantic access to simulator documentation, schemas, examples, and technical snippets. Procedural memory keeps high frequency simulator vocabulary and configuration patterns visible throughout the trajectory. A validator callable by the agent lets it check and repair candidate decks while drafting. A stop hook gated by validation prevents the agent from finishing with empty, incomplete, or structurally invalid outputs. Since SIGA exposes the simulator contract as a compact external adapter, it can also evolve from prior trajectories without modifying the underlying agent. Together, these components give the coding agent the missing executable contract while preserving its general software operation loop.

Our primary study uses GEOS, an open source multiphysics simulator for subsurface science. The main empirical finding is reliability. On easier validation tasks, the bare coding agent already operates near a quality ceiling. On harder held out GEOS tasks, SIGA improves mean TreeSim from 0.720 to 0.789 and reduces across run standard deviation by about 16 times by preventing catastrophic invalid outputs. A human calibration gives the practical scale of the bottleneck: on a representative GEOS deck authoring task, SIGA reaches in about five minutes the deck quality that a domain expert new to GEOS reached in about three hours.

We also test whether the adapter recipe transfers beyond GEOS. In OpenFOAM~\citep{weller1998tensorial}, the dominant bottleneck is structural completeness, so the most useful component is the completion gate that prevents missing required files. In LAMMPS~\citep{LAMMPS}, generated scripts are usually structurally complete, but the hard part is choosing correct values, unit conventions, command patterns, and domain specific parameters; there, procedural memory and retrieval carry the gain. This shift suggests a simple portable rule: use completion gates when whole blocks or files are missing, and use memory plus retrieval when value correctness is the bottleneck.

Our contributions are threefold. First, we formulate scientific simulator configuration as a practical bottleneck for AI for science and as an interface adaptation problem for coding agents. Second, we introduce SIGA, a lightweight and self improving adapter that supplies simulator grounding through retrieval, memory, validation, and termination control while leaving the base model and loop unchanged. Third, we show across GEOS, OpenFOAM, and LAMMPS that the same adapter family improves reliability, but the binding component depends on the simulator interface. These results suggest a practical route to automatic simulator configuration: build on today’s strongest coding agents, and adapt them to scientific software through small, portable, self-evolving interface adapters.

\section{Related work}
\label{sec:related}

\paragraph{LLM agents for scientific code and simulators.}
Scientific-coding agents are studied on curated benchmarks (ScienceAgentBench, DA-Code~\cite{chen2024scienceagentbench,huang2024dacode}) and built as general-purpose harnesses (OpenHands, SWE-agent~\cite{wang2024openhands,yang2024sweagent}), self-debugging loops~\cite{chen2024selfdebug}, or specialised search frameworks (SciNav~\cite{zhang2026scinav}). Domain-application systems include ChemCrow~\cite{bran2024chemcrow} and Coscientist~\cite{boiko2023coscientist} for chemistry, CellVoyager~\cite{alber2025cellvoyager} for single-cell analysis, and The AI Scientist~\cite{lu2024aiscientist,yamada2025ai} for end-to-end ML research. Closer to our setting are agents for scientific simulators: OpenFOAM-oriented computational fluid dynamics agents~\cite{chen2024metaopenfoam,pandey2025openfoamgpt,foamagent2025}, molecular-dynamics agents~\cite{shi2026mdagent2,guilbert2025dynamate,zhao2026polyjarvis}, finite-element/mechanics agents~\cite{zhan2025mooseagent,ni2024mechagents}, and reservoir-simulation agents~\cite{moyner2026jutulgpt}. The shared design lesson is that general-purpose LLMs are insufficient for high-fidelity simulator use unless wrapped with domain grounding, structured interfaces, execution feedback, and iterative correction. Unlike most of these we build \emph{on top of} an existing engineered coding harness (Claude Code) rather than writing an agent loop from scratch. Foam-Agent 2.0~\cite{foamagent2025} and MetaOpenFOAM~\cite{chen2024metaopenfoam} are the most direct comparators for our OpenFOAM transfer study (\S\ref{subsec:openfoam-transfer}); we compare against their lint-only execution mode and treat those as baselines.

\paragraph{Self-evolving agents, memory, and procedural guidance.}
Recent agent work treats capability as an iterative loop of planning, execution, feedback, and refinement that externalises experience into memory, search states, or revised plans~\citep{bran2024chemcrow,boiko2023coscientist,alber2025cellvoyager,lu2024aiscientist,zhang2026scinav,li2026agentflow,tang2026eigenagent,narayanan2024aviary}. A particularly active subfield, which treats the agent's own scaffolding as a learnable object (variously studied under the headings of meta-harness design~\cite{lee2026metaharness}, harness-as-code~\cite{ning2026codeagentharness}, agentic harness engineering~\cite{lin2026agenticharnessengineer}, and skill optimization~\cite{yang2026skillopt}), has shown promising results on core AI benchmarks (coding, terminal navigation, math).
Our self-evolved variant (\S\ref{subsec:self-evolved}) adopts this reflect-and-rewrite paradigm: the agent revises its own plugin, the adapter, based on prior trajectories.
Our focus is different: we study whether such self-revision helps on a task whose bottleneck is domain knowledge and procedural guidance rather than general programming competence. This connects to work on procedural memory and cheatsheets for LLMs, such as Buffer of Thoughts~\citep{yang2024bot}; we return
to this interface in \S\ref{sec:discussion}.

\section{Background: GEOS as a domain-specific simulator language}
\label{sec:background}

A GEOS~\citep{geos2024} deck is one or more XML files that specify a multiphysics simulation across ten canonical sections, covering the mesh, geometry, execution schedule, physics modules, material models, computational regions, numerical methods, field specifications, functions, and outputs. Although GEOS reads plain XML, its elaborate tag vocabulary and validation schema make its configuration language closer to a small domain-specific language (DSL); a deck is then a program in that language, whose tags name GEOS components and whose attributes set their parameters. For example, a physics-module tag selects which governing equations are advanced, a material-model tag defines the constitutive law, and region or output tags connect those choices to parts of the mesh and requested results. Writing a deck is therefore a translation from natural-language intent into a structured program over the simulator's executable interface. This is difficult for four reasons: (i) the vocabulary is large and contains near-duplicate component names, such as \texttt{DruckerPrager}, \texttt{ExtendedDruckerPrager}, \texttt{DruckerPragerHardening}, and \texttt{ViscoDruckerPrager}; (ii) documentation and examples may reflect older interface versions; (iii) constraints span sections, so names introduced in one place must be reused consistently elsewhere; and (iv) task briefs often under-specify choices that experts fill in with domain-conventional defaults. The first difficulty is exact interface-token selection, while the others are program-level consistency constraints. A concrete annotated deck is in App.~\ref{app:geos-example}.

\section{Method}
\label{sec:method}

\begin{figure}[t]
\centering
\includegraphics[width=\textwidth]{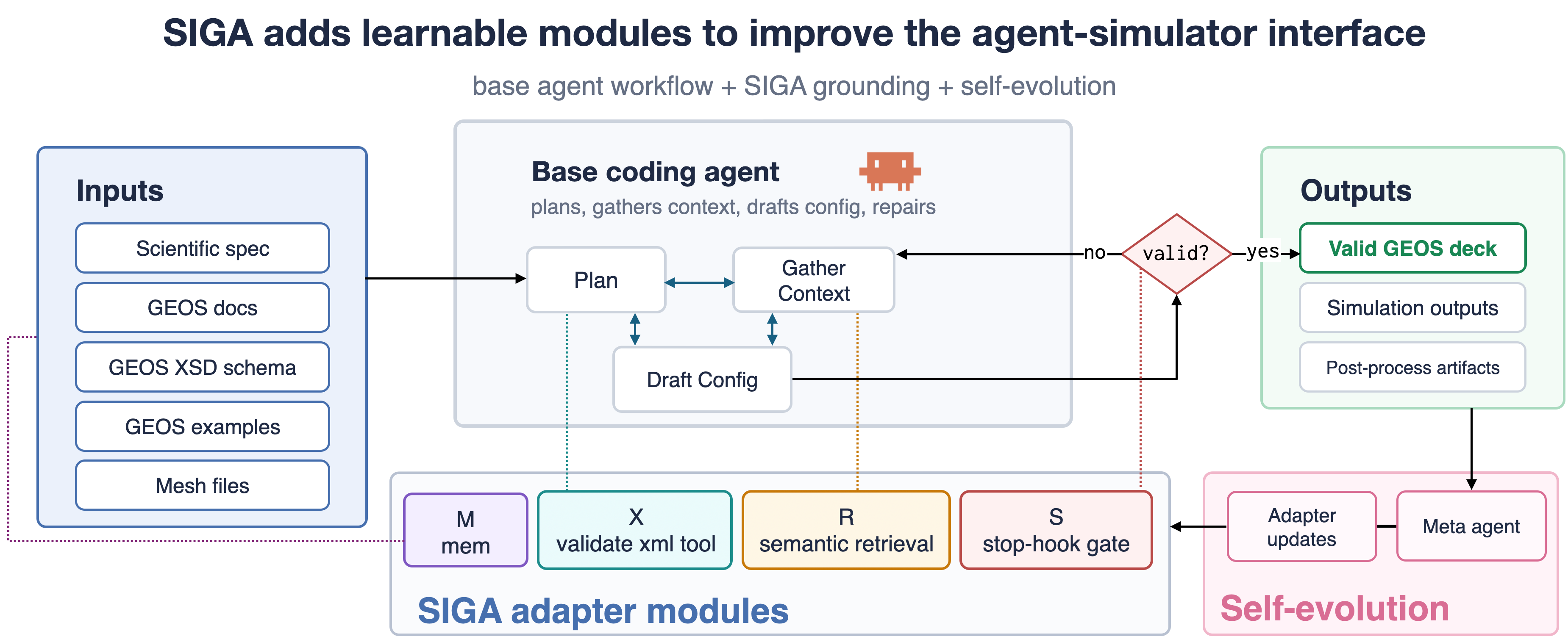}
\caption{\textbf{The SIGA method.} A natural-language simulation brief feeds into the base coding agent (a frozen harness $H_0$ wrapping a frozen model $\pi$), which runs its generic context\,$\to$\,act\,$\to$\,observe loop to author a configuration deck. The SIGA adapter grounds this loop at three interfaces, without modifying the loop itself: always-on \emph{procedural memory} (M) injected into the system context; \emph{retrieval} (R) and an \emph{XML validator} (X) added as agent-callable tools; and a schema-validating \emph{stop hook} (S) that gates termination, returning structured repair feedback until the deck parses and validates. The validated deck is emitted as output. The \emph{self-evolution} loop (dashed) reflects offline on logged trajectories to revise the adapter's contents (primer, memory, auxiliary skills), updating the adapter plugin package while the base harness $H_0$ and model $\pi$ stay fixed.}
\label{fig:2}
\end{figure}

\subsection{Overview}
\label{subsec:method-overview}

We frame simulator setup as an \emph{interface-grounding} problem for an off-the-shelf coding agent. A modern coding agent already supplies the generic machinery this task needs: it inspects files, drafts and edits candidate configurations, runs shell commands, reads back errors, and revises. What it lacks is the target simulator's \emph{executable contract}: the simulator-specific vocabulary, the structural and schema constraints a deck must satisfy, the means to check those constraints mid-edit, and the condition under which a workspace is complete enough to return. Setup failures therefore stem less from a missing agent loop than from missing interface grounding. SIGA supplies that grounding as a thin \emph{adapter} over a fixed coding harness, injected through the harness's own extension mechanisms rather than by rebuilding the loop (Fig.~\ref{fig:2}).

This adaptation-over-reconstruction stance is deliberate, for three reasons carried over from \S\ref{sec:intro} and \S\ref{sec:related}. (i) \emph{Portability}: porting to new scientific software requires rewriting only the simulator-facing contract, not a bespoke agent codebase. (ii) \emph{Optimization surface}: a small adapter is a tractable object for self-evolution (\S\ref{subsec:self-evolved}), whereas a sprawling agent loop exposes many opportunities for over-specification. (iii) \emph{Generality}: by grounding only the simulator-specific contract and leaving the generic loop untouched, the adapter does not overfit to the behavior of one model or one harness version. The same grounding layer should carry over to updated or swapped models and harnesses with little change, so that improvements in the underlying coding agent compound with the adapter rather than invalidating it, and adapting to new model or harness behavior stays cheap.

The grounding layer is also deliberately \emph{minimal}: rather than introduce new agent machinery, it instantiates three well-established ideas from the agent literature, each addressing a distinct simulator-setup failure. (1) \emph{Retrieval} (R) adds an alternate interface to domain knowledge through semantic query, for when the agent does not know the right simulator terms to search for. (2) \emph{Validator-driven self-refinement} supplies a correctness check that drives revision, in two complementary forms: agent-managed mid-trajectory (X) and externally enforced at termination (S). (3) \emph{Procedural memory} (M) retains useful experience across sessions by writing it down and keeping it in view. These four components plug into the harness at its context, tool, and termination interfaces, which we now formalize.

\subsection{The grounding adapter}
\label{subsec:overview}

\paragraph{Base harness and objective.} Fix a frozen base model $\pi$ wrapped by a frozen coding harness $H_0$. Given a natural-language task $x$ drawn from a task distribution $\mathcal{X}$ (a simulation brief), the harness executes a rollout $\tau \sim p_\pi(H_0, x)$: at each step it assembles a context from the system prompt, interaction history, and tool outputs; the model emits an action $a_t$ (a tool call or file edit); the environment returns an observation; the loop repeats until a termination predicate fires, leaving a workspace whose generated deck we denote $\hat{y}(\tau)$. A reward $r(\tau, x) = Q\big(\hat{y}(\tau),\, y^\star(x)\big) \in [0,1]$ scores the generated configuration against a hand-validated ground truth $y^\star(x)$ under the failures-as-zero convention (\S\ref{sec:eval}), where $Q$ is a task-appropriate quality metric: tree-edit similarity (TreeSim) for GEOS, defined in \S\ref{sec:eval}, and the file-coverage and LLM-judge metrics of \S\ref{subsec:openfoam-transfer} and \S\ref{subsec:lammps-transfer} for the transfer studies. Following \citet{lee2026metaharness}, the goal of any harness-level adaptation is to maximize expected reward over the harness rather than over model weights, $\max\, \mathbb{E}_{x\sim\mathcal{X}}\big[\,r(\tau,x)\,\big]$ with $\tau \sim p_\pi(H_0, x)$.

\paragraph{Three grounding interfaces.} Write the base harness as $H_0 = (c_0,\, \mathcal{T}_0,\, \mathrm{stop}_0)$: an initial system context $c_0$, a tool set $\mathcal{T}_0$, and a default termination predicate $\mathrm{stop}_0$. A SIGA adapter modifies exactly these three interfaces, and nothing else in the loop:
\begin{equation}
c_0 \;\mapsto\; c_0 \oplus m, \qquad
\mathcal{T}_0 \;\mapsto\; \mathcal{T}_0 \cup \mathcal{T}_{\mathrm{R}} \cup \mathcal{T}_{\mathrm{X}}, \qquad
\mathrm{stop}_0 \;\mapsto\; \mathrm{stop}_{\mathrm{S}},
\label{eq:adapter}
\end{equation}
yielding an adapted harness $H_A$ and rollouts $\tau \sim p_\pi(H_A, x)$. Localizing all grounding to these three well-defined points, context, tools, and termination, is what keeps the adapter thin and portable. The four components instantiate the three ideas above at these slots; each is binary (present or absent), and we describe them in turn, with implementation details in App.~\ref{app:impl}.

\paragraph{R: retrieval (tool interface).} The base agent may not reliably recover GEOS-specific tokens, such as solver names, material models, schema attributes, and example-compatible block combinations. The base coding agent can already retrieve over the simulator's documentation and example tree through its built-in \texttt{find}/\texttt{grep} tools, but this keyword search is brittle when the agent does not already know the right simulator terms to search for. \textbf{R} therefore adds semantic search over the same artifacts, the retrieval-augmented generation (RAG) approach that prior work has thoroughly investigated~\citep{lewis2021rag}: it extends the tool set with $\mathcal{T}_{\mathrm{R}}$, semantic access through a Model Context Protocol (MCP) server with three search tools: \texttt{search\_navigator} over GEOS documentation pages, \texttt{search\_schema} over XSD schema entries, and \texttt{search\_technical} over example XML files and technical snippets. It targets unknown-vocabulary substitution.

\paragraph{S, X: validator-driven self-refinement (termination and tool interfaces).} S and X expose the same schema check, \texttt{xmllint --schema} against the canonical GEOS \texttt{.xsd}, but at different control points. \textbf{S} acts at the termination interface, replacing $\mathrm{stop}_0$ with $\mathrm{stop}_{\mathrm{S}}$: on every attempted termination it scans \texttt{/workspace/inputs/} for parseable XML, validates it, and either allows the agent to finish or returns structured repair feedback, bounded by a per-task retry counter. S is mandatory whenever enabled and targets silent incompleteness, where the agent stops with an empty, unparseable, or schema-invalid deck. \textbf{X} acts at the tool interface, adding the same check to $\mathcal{T}_0$ as an optional MCP tool, \texttt{mcp\_\_xmllint\_\_validate\_geos\_xml}, that the agent may call while drafting, inspecting an intermediate file and revising before it finishes (about three calls per task on average when enabled); X targets in-trajectory schema-violation drift. Together they are the two faces of validator-driven self-refinement: externally enforced (S) and agent-managed (X).

\paragraph{M: procedural memory (context interface).} Some simulator knowledge is repeatedly useful across tasks and should not be rediscovered on every trajectory. \textbf{M} augments the always-on context, $c_0 \mapsto c_0 \oplus m$, where $m$ is a compact 775-token cheatsheet appended to the system prompt via \texttt{--append-system-prompt}, containing GEOS physics-module families, XML element names, constitutive-model names, typical attributes, and short exemplars, distilled offline from 18 training trajectories using a model different from the inference model. It is not an explicit policy but a lightweight memory surface that keeps high-frequency simulator conventions in view, and targets recurring-vocabulary lookup and repeated domain-interface mistakes.

\paragraph{The SIGA design space.} A \textbf{Simulator-Interface Grounding Adapter (SIGA)} is a plugin built on an existing coding agent from a small, fixed set of grounding ideas: cross-session procedural memory (M), semantic retrieval over domain knowledge (R), and validator-driven self-refinement applied mid-trajectory (X) and enforced at termination (S). Each idea plugs into the harness at one of its context, tool, or termination interfaces. SIGA is therefore not a single fixed configuration but a compact design space over these four ideas. A concrete adapter is a point $b \in \{0,1\}^{\{\mathrm{R,S,X,M}\}}$ in this space, selecting which ideas are enabled, with $b = \mathbf{0}$ recovering the bare harness (\emph{Vanilla}). Adapting to a new simulator then means instantiating the relevant ideas against that simulator's contract and keeping those that help, since which idea is binding depends on the interface (\S\ref{sec:eval}); in this paper we map the space by factorial ablation (\S\ref{subsec:cells}). This small, bounded design space, rather than a bespoke agent loop, is what makes the adapter cheap to port and tractable to self-evolve.

\subsection{Self-evolving the adapter}
\label{subsec:self-evolved}

The slots above fix \emph{where} grounding enters; their textual contents are additional free parameters. Let $\theta$ collect the adapter's contents, the system primer, the memory cheatsheet $m$, and any agent-authored auxiliary skills, giving a parameterized adapter $A_\theta$. Hand-design fixes $\theta$ in advance; we instead test a modest form of self-evolution that searches for it,
\begin{equation}
\theta^\star \;=\; \arg\max_{\theta}\; \mathbb{E}_{x\sim\mathcal{X}_{\mathrm{sel}}}\big[\,r(\tau,x)\,\big], \qquad \tau \sim p_\pi\big(H_{A_\theta},\, x\big),
\label{eq:selfevolve}
\end{equation}
on a held-out validation-selection split $\mathcal{X}_{\mathrm{sel}}$. Mirroring the meta-harness outer loop \citep{lee2026metaharness}, the search is run offline by a coding-agent proposer that reads prior candidates' contents, trajectories, and rewards from a filesystem and proposes revised contents, retaining those that improve $\mathcal{X}_{\mathrm{sel}}$ reward. Crucially, this does not modify the underlying coding agent, introduce a new planning policy, or search the full agent loop: the base harness $H_0$ and model $\pi$ stay frozen, and only the adapter contents $\theta$ move, a far smaller search space than the entire harness program. This lets us ask whether the adapter can improve from its own execution history on a domain-knowledge-heavy task, whereas prior self-improving-agent and harness-optimization work mostly targets general coding, terminal, or reasoning benchmarks. We evaluate two variants to separate gains from improved prose grounding versus the broader self-evolved package.

\paragraph{SE: self-evolved combined adapter.} SE is produced by the offline pipeline that iteratively rewrites the adapter contents, including a richer task primer (system prompt element) and agent-authored auxiliary skills, scoring candidate revisions by resulting quality on a held-out training-set. At evaluation time we disable skill invocation with \texttt{--disallowedTools Skill} so that SE has the same tool shape as the factorial cells; any gain in the main comparison therefore reflects the evolved adapter contents rather than an additional class of runtime tools.

\paragraph{SE-prose: prose-only self-evolved variant.} SE-prose isolates the textual part of the self-evolved adapter: it takes only the rewritten primer and cheatsheet from SE and inserts them into an otherwise standard S+X+M cell. Comparing SE-prose with S+X+M tests whether self-evolution improves the always-visible grounding instructions and procedural memory; comparing SE-prose with SE tests whether any remaining gain comes from the broader self-evolved package beyond prose and memory.


\section{Experiments}
\label{sec:eval}







\subsection{Experimental Design}

Our experiments are designed to separate three questions that are often conflated in simulator-agent evaluation: 
whether the adapter improves final deck quality, whether it improves reliability by preventing catastrophic invalid 
outputs, and which simulator-interface failures remain unsolved. We first run a controlled factorial ablation on GEOS 
to estimate the effect of each grounding component. We then analyze failure categories to identify the residual 
bottlenecks. Finally, we calibrate the task against human domain users, test behavior under underspecified briefs,
and run two cross-simulator transfer studies (OpenFOAM and LAMMPS) to probe whether the dominant mechanism is GEOS-specific.

The results answer five research questions:
\textbf{RQ1: Reliability and quality.} Does SIGA improve TreeSim, and is the gain driven by average quality or by fewer catastrophic failures?
\textbf{RQ2: Failure mechanisms.} Which block-level or attribute-level errors are fixed by the adapter?
\textbf{RQ3: Human anchor.} How does SIGA compare with domain experts on a representative GEOS deck-authoring task?
\textbf{RQ4: Autonomy.} When task briefs are underspecified, does the agent consult a human supervisor or substitute other available information sources?
\textbf{RQ5: Transfer.} Does the same grounding mechanism transfer to other simulator interfaces (OpenFOAM, LAMMPS), and does the dominant component shift with the interface?






\subsection{Ablation design and bottleneck analysis}
\label{subsec:cells}

\textbf{Cells.} To attribute the contribution of each grounding component, we run a detailed factorial ablation across the four binary factors $\{\mathrm{R, S, X, M}\}$. A standard one-factor-at-a-time ablation (stacking each component on, or stripping it from the full stack) is the cheapest design, but it cannot disambiguate main effects from two-factor interactions: if R and S help only when combined, neither single-factor ablation reveals this. The full $2^4 = 16$-cell factorial does, at the cost of substantially more runs. We instead use a \emph{Resolution-IV $2^{4-1}$ fraction} with generator $\mathrm{D}{=}\mathrm{ABC}$, which gives us eight cells whose main effects are not confounded with two-factor interactions, recovering most of the information of the full factorial at half the compute. Cells are named by their active factors, e.g., $\mathrm{X+M}$; \emph{Vanilla} is all-off. We add three cells: $\mathrm{S+X+M}$, the predicted main-effects-best corner missing from the fraction; \emph{SE-prose}, which inserts the self-evolved v3 primer and cheatsheet into an otherwise standard $\mathrm{S+X+M}$ cell; and \emph{SE}, the full self-evolved combined adapter. Because S and X both use \texttt{xmllint}, the X main effect partly conflates agent-callable validation with hook-time schema validation when S is also enabled (App.~\ref{app:discussion}).

\textbf{Bottleneck analysis.} We attribute score differences to structural failure modes in three steps (schema and prompts in App.~\ref{app:bottleneck}). First, an LLM-free extractor identifies top-$K$ failing subtrees from \texttt{treesim\_detail} and adds trajectory features from \texttt{events.jsonl}. Second, \texttt{deepseek-v4-flash} assigns each (cell, run, task) to one category: \texttt{missing\_block}, \texttt{extra\_block}, \texttt{hallucinated\_extras}, \texttt{structural\_mismatch}, \texttt{bad\_attribute\_value}, \texttt{partial\_implementation}, \texttt{wrong\_constitutive}, or \texttt{no\_failure}. Third, \texttt{deepseek-v4-pro} summarizes per-cell distributions and baseline-vs-best deltas. Total cost is ${\sim}650$ flash calls plus 4 pro calls ($\$5\text{--}7$).

\textbf{Splits and leakage control.} From the 46-task pool we reserve 10 tasks for held-out evaluation, 18 for distillation, and 17 for validation-selection, dropping one task (App.~\ref{app:bench}). No validation or held-out-eval task is used for distillation. A hygiene gate regex-scans distilled artifacts for ground-truth test filenames and rejects leaks. (It was added after an earlier cheatsheet leaked 13/17 validation-task basenames.)

\subsection{GEOS benchmark and evaluation protocol}

We evaluate on two GEOS task splits. The validation (val) split contains 17 in-distribution tasks from advanced examples 
and tutorials, covering poromechanics, hydraulic fracture, thermal coupling, and wellbore modeling. This split is used 
for component selection but not for memory distillation. The held-out-eval split contains 10 harder tasks reserved 
for final evaluation and is never used for distillation, self-evolution, or cell tuning.

Generated decks are scored with TreeSim, a recursive tree-similarity metric in $[0,1]$ over the parsed XML deck
(full definition and scorer in App.~\ref{app:treesim}). Each XML element is a node labeled by its tag and \texttt{name}
attribute; \texttt{<Included>} directives are resolved and a deck's files are merged under a synthetic root, so file-level
(single file) and deck-level (full directory) scores use the same procedure. Same-tag children are matched by greedy
\emph{unordered} bipartite matching on a tag-plus-attribute similarity, so sibling order does not affect the score. A
node's score blends its own attribute similarity $a$ (a Jaccard-style overlap of attribute keys, values compared by
case-insensitive string match or relative numeric tolerance) with the mean score of its matched children,
\begin{equation}
s \;=\; \alpha\, a \;+\; (1-\alpha)\,\bar{s}_{\mathrm{child}} \;-\; \beta\,\frac{n_{\mathrm{extra}}}{n_{\mathrm{gt}}+n_{\mathrm{extra}}},
\qquad \alpha=0.3,\ \beta=0.1,
\label{eq:treesim}
\end{equation}
clamped to $[0,1]$, where leaves use $s=a$, unmatched ground-truth children contribute $0$ to $\bar{s}_{\mathrm{child}}$,
and $n_{\mathrm{extra}}$ surplus generated children incur the penalty term. The deck score is the root node's score;
per-section scores are the scores of the ten top-level children. We report failures-as-zero: parse errors, timeouts,
missing XML outputs, and empty outputs all receive score 0. This convention is important because simulator setup is only
useful when the returned workspace is at least structurally inspectable.

The main baseline is vanilla Claude Code with all SIGA components disabled. All headline GEOS experiments use 
\texttt{deepseek-v4-flash} with three runs per cell and a 1500-second timeout. SE and SE-prose use the same evaluation budget 
and disable skill invocation for parity.


\section{Results}
\label{sec:results}

The main empirical finding is that simulator-interface grounding's largest gain is reliability. On in-distribution GEOS
tasks, vanilla Claude Code already achieves high TreeSim. On harder held-out tasks, SIGA reduces catastrophic invalid or
incomplete outputs, which raises mean performance and sharply reduces across-run variance. We further localize the
residual errors to attribute-level semantic mistakes (\S\ref{subsec:bottleneck-results}), which marks where future
grounding can extend the gains.

\subsection{Reliability is SIGA's biggest gain (RQ1)}
\label{subsec:quality}

Table~\ref{tab:main-results} reports headline numbers ($n=3$). Cells cluster narrowly on val (0.857 to 0.921; 0.910 to 0.921 for sans-semantic-retrieval) but spread more on held-out-eval (0.720 to 0.789, $\Delta = +0.069$ Vanilla to SE). The gap between these two columns, more than the absolute level of either, is the headline finding: the adapters' contribution is largely reliability rather than uniform quality lift.

\begin{table}[t]
  \caption{Cell-level TreeSim (failures-as-zero) on \texttt{deepseek-v4-flash}, $n=3$ runs (mean $\pm$ sample std). The left block indicates which grounding components are present in each cell (\textbf{$\bullet$} on, blank off): \emph{Retrieval} (semantic search over GEOS artifacts), \emph{Refine loop} (schema-validating termination hook), \emph{Validator} (the same schema check exposed as an agent-callable tool), and \emph{Memory} (procedural-memory cheatsheet appended to the system prompt). The first eight rows are a Resolution-IV $2^{4-1}$ fractional factorial; \emph{S+X+M} fills the main-effects-best corner missing from the fraction; \emph{Self-Evolve-prose} substitutes the self-evolved primer and cheatsheet into an otherwise standard \emph{S+X+M} cell; \emph{Self-Evolve} is the full self-evolved adapter (the two are abbreviated SE-prose and SE elsewhere). $\Delta$ = cell mean minus \emph{Vanilla} mean on the same split.}
  \label{tab:main-results}
  \centering
  \small
  \resizebox{0.98\textwidth}{!}{%
  \rowcolors{3}{gray!12}{white}
  \begin{tabular}{l|cccc|cccc}
    \toprule
    & \multicolumn{4}{c|}{\textbf{Components}} & \multicolumn{4}{c}{\textbf{TreeSim}} \\
    \textbf{Cell} & \textbf{Retrieval} & \textbf{Refine loop} & \textbf{Validator} & \textbf{Memory} & \textbf{val} & \textbf{held-out-eval} & \textbf{$\Delta$ val} & \textbf{$\Delta$ held-out} \\
    \midrule
    Vanilla        &         &         &         &         & $0.910 \pm 0.024$           & $0.720 \pm 0.081$            & --                & --                \\
    R$+$M          & $\bullet$ &         &         & $\bullet$ & $0.885 \pm 0.014$           & --                           & $-0.025$          & --                \\
    S$+$M          &         & $\bullet$ &         & $\bullet$ & $0.919 \pm 0.004$           & --                           & $+0.009$          & --                \\
    R$+$S          & $\bullet$ & $\bullet$ &         &         & $0.857 \pm 0.045$           & --                           & $-0.053$          & --                \\
    X$+$M          &         &         & $\bullet$ & $\bullet$ & $\mathbf{0.921 \pm 0.007}$  & $0.768 \pm 0.005$            & $\mathbf{+0.011}$ & $+0.048$          \\
    R$+$X          & $\bullet$ &         & $\bullet$ &         & $0.893 \pm 0.033$           & --                           & $-0.017$          & --                \\
    S$+$X          &         & $\bullet$ & $\bullet$ &         & $0.917 \pm 0.004$           & $0.781 \pm \mathbf{0.002}$   & $+0.007$          & $+0.061$          \\
    R$+$S$+$X$+$M  & $\bullet$ & $\bullet$ & $\bullet$ & $\bullet$ & $0.885 \pm 0.008$           & --                           & $-0.025$          & --                \\
    \midrule
    S$+$X$+$M      &         & $\bullet$ & $\bullet$ & $\bullet$ & $0.911 \pm 0.018$           & $0.783 \pm 0.022$            & $+0.001$          & $+0.063$          \\
    Self-Evolve-prose &      & $\bullet$ & $\bullet$ & $\bullet$ & $0.897 \pm 0.032$           & $0.775 \pm 0.024$            & $-0.013$          & $+0.055$          \\
    Self-Evolve    &         & $\bullet$ & $\bullet$ & $\bullet$ & $0.919 \pm 0.020$           & $\mathbf{0.789 \pm 0.012}$   & $+0.009$          & $\mathbf{+0.069}$ \\
    \bottomrule
  \end{tabular}}
\end{table}

\textbf{Hard-tail rescue.} The bare harness's high across-run variance on held-out-eval comes from a small number of catastrophic outputs, not from broad spread: the agent occasionally terminates with an empty or unparseable deck that scores 0, and a single such run inflates the standard deviation for the whole cell. The adapters' first-order job is to prevent these zero-score terminations, which is why they cut across-run standard deviation by roughly an order of magnitude (e.g.\ Vanilla $\sigma = 0.081$, driven by one unparseable \texttt{ExampleProppantTest} run, falls to $\sigma = 0.005$ under X+M and $\sigma = 0.002$ under S+X) and lift the mean on the harder split by rescuing the would-be zero-score runs. Per-task inspection (Table~\ref{tab:per-task-icl10}) shows the $+0.069$ Vanilla-to-SE aggregate gain on held-out-eval is driven primarily by two catastrophic-failure rescues (\texttt{AdvancedExampleThermoPoroElasticWellbore} from $0.355$ to $0.761$; \texttt{ExampleProppantTest} from $0.541$ to $0.825$), alongside one universal failure (\texttt{TutorialHydraulicFractureWithAdvancedXML} at $0.013$ across every cell) and within-noise differences on the remaining tasks. The adapters rescue the bare model from these catastrophic failures rather than uniformly raising scores: where the bare model already has a usable template, the adapter is operating in run-to-run noise.

\label{subsec:per-task}\textbf{Val ceiling.} On the validation split (val), where the bare harness already operates near a quality ceiling, the absolute-score improvements from any single component are subtle. The best val cell (X+M) beats Vanilla by $+0.011$, within run-to-run variation; this is a modest absolute lift, and the headline contribution on val is again reliability (variance across runs) and efficiency (\S\ref{subsec:efficiency}) rather than absolute quality. The largest Resolution-IV main effect on val is R ($\Delta = -0.037$); the other three are small ($\mathrm{X}\,{+}0.011$, $\mathrm{M}\,{+}0.008$, $\mathrm{S}\,{-}0.008$), all within $\pm 0.011$ of zero.

\begin{rqanswer}{RQ1}
SIGA's largest gain is reliability: on held-out-eval it shrinks across-run variance by roughly an order of magnitude and rescues two compound multi-physics tasks from catastrophic failure. On the in-distribution split the bare harness already sits near a quality ceiling, so the absolute-quality lift there is small.
\end{rqanswer}

\subsection{Schema-aware adapters fix block-level omissions (RQ2)}
\label{subsec:bottleneck-results}

The per-task error classifier (App.~\ref{app:bottleneck-counts}) reveals four patterns. (1) Adapters fix \texttt{missing\_block}: counts drop 6 to 3 on val between Vanilla and X+M (whole \texttt{Solvers}, \texttt{Events}, or \texttt{Constitutive} omissions); the cheatsheet enumerates these canonical blocks and the validator's structural pressure discourages finishing without them. (2) Adapters do not fully resolve \texttt{bad\_attribute\_value} (12 / 11 / 15 across Vanilla / X+M / S+X): schema validation cannot prevent \texttt{TPFAstabilization} substituted for \texttt{contactStabilization} or a hallucinated \texttt{gravityVector}. (3) Adapters trade catastrophic absence for content imprecision: as \texttt{missing\_block} drops, \texttt{extra\_block} rises 9 to 11 and \texttt{hallucinated\_extras} rises 4 to 7, yielding small mean lift but large reliability lift. (4) Strictly perfect tasks (TreeSim $\geq 0.999$) do not increase under any adapter (Vanilla 7/51, X+M 6/51, SE 6/51).

Two points put the persistent \texttt{bad\_attribute\_value} count in context. First, choosing the right attribute value is intrinsically harder than including the right block: the correct value often follows from domain knowledge that the schema and the shipped documentation do not encode, such as a physically appropriate solver tolerance, permeability, or constitutive choice. Closing this gap is therefore less a matter of a stricter validator than of supplying the relevant domain sources (e.g.\ from primary literature) or adding scientific-reasoning capability, both of which are engineering directions beyond the present grounding layer. Second, the roughly flat count is partly mechanical: as the adapters fill in previously-missing blocks, more attributes are now present to be judged at all, so a near-constant \texttt{bad\_attribute\_value} count alongside falling \texttt{missing\_block} reflects more attributes being inferred rather than a regression. The adapters thus remove the most egregious, whole-block errors and leave a relatively more minor, attribute-level residue, which we read as the next problem to solve rather than a shortcoming of the approach.

\paragraph{Efficiency.}\label{subsec:efficiency} Adapter cells are no more expensive than Vanilla in wall-clock (App.~\ref{app:results}). This matched-efficiency outcome is itself a feature rather than a null finding. A common failure mode in LLM-reflection-driven harness optimization is over-specification, where iteratively tacked-on small features end up derailing or slowing down workflows, so a non-trivial design constraint is that the adapter not impose runtime overhead beyond the bare harness. Our adapters meet this constraint: cheatsheet cells make about half as many free-form \texttt{Read} calls because the cheatsheet short-circuits exploratory file access; xmllint cells make roughly 3 schema-validation calls per task; RAG cells make 12 to 13 retrieval calls and run faster but score lower. SE makes fewer tool calls than Vanilla on val (68.9 vs 81.5, ${-}15.5\%$) but more on held-out-eval (97.4 vs 90.5, $+7.6\%$); the val efficiency does not transfer to the harder task split, consistent with the broader pattern that adapter wins are tail-localised.

\begin{rqanswer}{RQ2}
Schema-aware adapters reliably fix block-level omissions and add no wall-clock overhead. The residual errors are attribute-level: assigning physically correct values is a harder, domain-knowledge-bound problem that the schema does not constrain, and it is a less severe error class than whole-block omission. We see it as the next frontier for grounding rather than a limitation of the recipe.
\end{rqanswer}

\subsection{Human baseline: SIGA reaches expert deck quality faster (RQ3)}
\label{subsec:human-baseline}

To anchor agent TreeSim in human terms, two geoscience-domain-expert volunteers (Expert 1, Expert 2; graduate-level geoscientists, new to GEOS) attempted \texttt{buckleyLeverettProblem} (1D immiscible CO$_2$/brine displacement, the easy end of our bench) in a single one-hour session, using the same primer, filtered source tree, and contamination block as the agent and working primarily from the GEOS documentation and source tree. Expert 1 returned with no time cap as an extended-budget check, and we collected a written estimate from a GEOS expert and developer to anchor experienced users (App.~\ref{app:human-browser}).

Each one-hour session opened with roughly ten minutes of task explanation and environment setup; in the authoring time that remained, both experts completed only \texttt{base.xml} (the first of the two required files), spending the latter part of the session on the \texttt{Outputs} and \texttt{Events} blocks. File-level TreeSim is $0.812$ (Expert 1) and $0.781$ (Expert 2); deck-level collapses to $0.540$ and $0.527$ since the second file is missing (Table~\ref{tab:human-baseline}). On the same task, vanilla Claude Code reaches file-level $0.889 \pm 0.023$ and deck-level $0.751 \pm 0.016$ in roughly 7 min; SIGA X+M reaches at least $0.90$ on both. The single-file outcome reflects the one-hour budget, not capacity: Expert 1, in an uncapped follow-up session, produced both files in about 3 hours and reached deck-level $\mathbf{0.931}$, at parity with SIGA X+M but at roughly $36\times$ the wall-clock. The GEOS developer estimate brackets an experienced GEOS user at under 30 min for simple Buckley--Leverett and ``a couple of days'' for compound multi-physics decks, lining up with the hard-tail held-out-eval result.

\begin{table}[h]
  \caption{Human baseline on \texttt{buckleyLeverettProblem}. \emph{Quality} columns are TreeSim (higher is better, $\uparrow$): file-level is GT \texttt{base.xml} vs the participant's \texttt{base.xml}; deck-level is the full GT directory vs the submitted directory. The \emph{Efficiency} column is wall-clock minutes to a final deck (lower is better, $\downarrow$). Agent rows from per-task \texttt{deepseek-v4-flash} numbers.}
  \label{tab:human-baseline}
  \centering
  \small
  \begin{tabular}{lrrr}
    \toprule
    & \multicolumn{2}{c}{\textbf{Quality} ($\uparrow$ better)} & \textbf{Efficiency} ($\downarrow$ better) \\
    \cmidrule(lr){2-3}\cmidrule(lr){4-4}
    \textbf{Author} & \textbf{File-level (base)}\,$\uparrow$ & \textbf{Deck-level}\,$\uparrow$ & \textbf{Wall (min)}\,$\downarrow$ \\
    \midrule
    Expert 1 (1\,h session)                         & $0.812$            & $0.540$            & $48.2$ \\
    Expert 2 (1\,h session)                         & $0.781$            & $0.527$            & $46.7$ \\
    Expert 1 (no time cap, both files)              & $0.689$            & $0.931$            & ${\sim}180$ \\
    Vanilla CC (\texttt{deepseek-v4-flash})   & $0.889 \pm 0.023$  & $0.751 \pm 0.016$  & ${\approx}\,7$ \\
    SIGA X+M (\texttt{deepseek-v4-flash})     & ${\geq}\,0.90$     & ${\geq}\,0.90$     & ${\approx}\,5$ \\
    \bottomrule
  \end{tabular}
\end{table}

Browser histories show humans browsing Sphinx prose to assemble decks from concept descriptions, while the agent analogises from prior decks via the source tree (App.~\ref{app:human-browser}). Part of the SIGA contribution is giving the agent a structured way to do what humans approximate via Sphinx browsing under time pressure. We treat this as an existence-of-effect calibration, not a head-to-head ranking.

\begin{rqanswer}{RQ3}
The SIGA-equipped agent reaches the deck-level quality of an extended-budget domain expert on this representative task at roughly $1/36$ the wall-clock, and operates inside the experienced-GEOS-user time envelope for simple Buckley--Leverett deck authoring.
\end{rqanswer}

\subsection{Exploring agent autonomy: human consultation is under-utilized in hard settings (RQ4)}
\label{subsec:autonomy}

\textbf{Motivation.} Our headline benchmark hands the agent a fully detailed task brief in a single turn, by design: this standardised setup removes potential ambiguity and confounders, so any difference between adapter cells is attributable to the adapter rather than to varying levels of underspecification. In this regime the agent's responsibility is narrowed to translation: it is told exactly which simulation to run, and only needs to express it in GEOS's DSL. Eventually, however, we want to expand the scope of the agent's responsibility so that it can take on more of the workflow and be correspondingly more useful to a working scientist. Loosening the brief naturally introduces ambiguity that a human scientist collaborator would otherwise resolve. We explore this direction by producing alternative specifications that specify less and less of the task, requiring the agent to figure out more on its own, and observing when and how the agent chooses to consult a human expert user to resolve the ambiguities it encounters.

\textbf{Protocol.} We probe this with a single-run companion study (8 val tasks, 2 configurations $\times$ 2 relaxation levels $\times$ 2 interaction modes, 64 runs; App.~\ref{app:autonomy}). Briefs are tier-rewritten by \texttt{deepseek-v4-pro}: \textbf{Medium} drops T1 (software defaults) and T2 (standard numerics), 89 values ($-21\%$ chars); \textbf{Hard} additionally drops T3 (domain-inferable physicals), 184 values ($-50\%$). Configurations are Vanilla and X+M; the \emph{interactive} mode exposes a \texttt{consult\_supervisor} MCP tool whose handler invokes a separate \texttt{deepseek-v4-flash} prompted with the full original brief.

\textbf{Results.} Across 64 interactive trials the agent invoked the exposed human-consultation tool only twice ($3.1\%$), a rate that is robust to prompt framing. The 31/32 silent runs were not silent for lack of questions: agents read 142 to 404 GEOS example XMLs per cell and copied values from analogous benchmarks. A neutral-framing rerun (no ``prefer to infer'' language) leaves the rate unchanged at 1/32, ruling out prompt-induced reluctance. A Mandel/Hard diagnostic shows 15 of 26 dropped values appear by literal-token \texttt{grep} in other readable GEOS examples, so the on-disk library is acting as a cheaper retrieval substitute for human consultation. TreeSim drops mildly with relaxation (X+M: $0.829$ Medium, $0.835$ Hard, vs $0.921$ at Easy), and interactive variants are within $\pm 1$pp of non-interactive: the agent reaches the same place whether or not the human channel exists.

\begin{rqanswer}{RQ4}
``Decide when to consult human oversight'' is not a free-standing agent capability in this setting; it is conditional on whether the agent has access to an alternative oracle. The on-disk example library acts as a cheaper retrieval substitute for human consultation, so studies measuring consultation behaviour without controlling for executable-example-library access are measuring the substitute, not the consultation. This is also a benchmark-design lesson: eliciting genuine consultation requires tasks whose missing information cannot be recovered from accessible examples, so that asking a human is the only reliable path. Our relaxed briefs did not clear that bar, the agent judged them close enough to examples it could already read, so designing tasks that force a real ambiguity is itself an open methodological problem.
\end{rqanswer}

\subsection{Transfer to OpenFOAM: SIGA produces complete cases where native agents fail (RQ5)}
\label{subsec:openfoam-transfer}

We first ask whether the dominant SIGA mechanism is specific to GEOS's XML/XSD interface or reflects a broader adapter principle: forcing the agent to satisfy simulator-specific completion checks before termination. We port $\{\mathrm{R,S,X,M}\}$ to OpenFOAM case authoring and run a 30-task study drawn from a FoamGPT-derived benchmark, using a single run, \texttt{deepseek-v4-flash}, and a file-text-and-coverage metric ($0.7\cdot\mathrm{mean\,similarity} + 0.3\cdot\mathrm{coverage}$; App.~\ref{app:openfoam-transfer}). To place the result against simulator-native systems, we additionally run two from-scratch OpenFOAM agents, Foam-Agent~\citep{foamagent2025} and MetaOpenFOAM~\citep{chen2024metaopenfoam}, both in lint-only execution mode (the stable mode available in our environment), under the same benchmark and metric with matched token instrumentation.

The pattern matches the GEOS reliability result and now stands out against external baselines (Table~\ref{tab:openfoam-main}). The best OpenFOAM cell, R+S, reaches mean $0.870$, and \emph{every} SIGA cell, including Vanilla, produces all required files on all 30 tasks with no zero-score outputs. The two native agents fail exactly there: Foam-Agent reaches $0.516$ (19/30 full coverage, 8 zero-score) and MetaOpenFOAM $0.379$ (10/30 full coverage, 12 zero-score), and their score collapses are dominated by missing-required-file failures rather than textual mismatch. Among SIGA cells the stop-hook carries the quality gain: the $\mathrm{S}$ factor appears in all five top cells, and a factor-style readout assigns $\mathrm{S}$ the largest positive effect ($+0.168$ mean) while $\mathrm{R}$ ($+0.005$), $\mathrm{X}$ ($+0.007$), and $\mathrm{M}$ ($-0.007$) are near zero (App.~\ref{app:openfoam-transfer}). The harness buys this coverage with cost: the SIGA cells run roughly an order of magnitude more expensive per task than the native agents, which resend less context per turn.

\begin{table}[t]
  \caption{OpenFOAM transfer study on the 30-task \texttt{n30\_hybrid} benchmark (\texttt{deepseek-v4-flash}, single run). Mean score is the file-text-and-coverage metric. \emph{Full cov} = tasks with all required files produced; \emph{Zero} = zero-score tasks. Eight rows form the Resolution-IV $2^{4-1}$ fraction and S+X+M is an additional hand-selected cell (nine SIGA cells total); the last two are OpenFOAM-native agents in lint-only mode. Est.\ cost is the per-row total over all 30 tasks, recomputed from logged input/output tokens at OpenRouter DeepSeek V4 Flash pricing.}
  \label{tab:openfoam-main}
  \centering
  \small
  \resizebox{0.98\textwidth}{!}{%
  \begin{tabular}{lrrrrr}
    \toprule
    \textbf{System} & \textbf{Mean score} $\uparrow$ & \textbf{Full cov} $\uparrow$ & \textbf{Zero} $\downarrow$ & \textbf{Wall s} & \textbf{Est.\ cost (\$)} \\
    \midrule
    SIGA R$+$S            & $\mathbf{0.870}$ & 30/30 & 0/30 & 263.5 & 2.61 \\
    SIGA S$+$X            & $0.866$ & 30/30 & 0/30 & 469.8 & 3.68 \\
    SIGA R$+$S$+$X$+$M    & $0.854$ & 30/30 & 0/30 & 376.9 & 3.81 \\
    SIGA S$+$X$+$M        & $0.847$ & 30/30 & 0/30 & 495.4 & 3.65 \\
    SIGA S$+$M            & $0.835$ & 30/30 & 0/30 & 429.5 & 3.29 \\
    SIGA X$+$M            & $0.697$ & 30/30 & 0/30 & 236.0 & 2.04 \\
    SIGA R$+$M            & $0.689$ & 30/30 & 0/30 & 220.8 & 1.15 \\
    SIGA R$+$X            & $0.685$ & 30/30 & 0/30 & 179.2 & 1.62 \\
    SIGA Vanilla          & $0.681$ & 30/30 & 0/30 & 196.3 & 1.36 \\
    \midrule
    Foam-Agent 2.0 (lint) & $0.516$ & 19/30 & 8/30  & 373.0 & 0.30 \\
    MetaOpenFOAM (lint)   & $0.379$ & 10/30 & 12/30 & 431.5 & 0.17 \\
    \bottomrule
  \end{tabular}}
\end{table}

\begin{rqanswer}{RQ5 (OpenFOAM)}
The dominant GEOS mechanism transfers: forced end-of-turn verification ($\mathrm{S}$) is again the strongest component, and across 30 tasks the SIGA harness returns complete, structurally valid cases on every task while two OpenFOAM-native agents, in our lint-only reproduction, leave 8--12 tasks with missing required files. The transferable principle is the completeness gate, instantiated as XSD validity in GEOS and required-file/dictionary coverage in OpenFOAM.
\end{rqanswer}

\subsection{Transfer to LAMMPS: the dominant component shifts to knowledge injection (RQ5)}
\label{subsec:lammps-transfer}

We next port $\{\mathrm{R,S,X,M}\}$ to LAMMPS molecular-dynamics input-script authoring on a 9-task benchmark spanning diverse MD physics (single run, LLM-judge metric in $[0,10]$; App.~\ref{app:lammps-transfer}). To probe transfer across both simulator and model family, we run 12 configurations across two backbone models, Claude Sonnet 4.6 and \texttt{deepseek-v4-flash} (both inside the Claude Code harness). The recipe transfers, but with a shifted factor profile (Table~\ref{tab:lammps-headline}). On DeepSeek the full stack reaches mean $7.78$ versus $4.56$ for the bare agent, a $3.2$-point gain; on Claude the best cell reaches $6.89$ versus $6.33$, a smaller margin because the bare Claude agent already assigns mostly correct parameter values. A factor-style readout on DeepSeek assigns the largest effect to $\mathrm{M}$ ($+2.13$), then $\mathrm{R}$ ($+1.55$), with $\mathrm{S}$ and $\mathrm{X}$ positive but smaller; on Claude all effects are small (at most $0.52$ in magnitude). Unlike GEOS and OpenFOAM, where forced verification ($\mathrm{S}$) dominated, the LAMMPS gain is informational, supplied by memory and retrieval: correct unit conventions, ensemble syntax, and task-specific command patterns. Structural scores are near ceiling across all 12 configurations ($\geq 0.976$), confirming that the agents reliably emit complete scripts and that the residual errors are wrong parameter values rather than omitted components, which is why knowledge injection rather than completeness enforcement carries the signal here.

\begin{table}[t]
  \caption{LAMMPS transfer headline (9 tasks, single run, LLM-judge mean in $[0,10]$). Two backbones; the best SIGA configuration is the one with the highest mean for that backbone (for Claude, M+R and M+R+S+X tie at $6.89$; we list M+R). Structural scores are near-ceiling ($\geq 0.976$) for all cells, so the differentiation is in judged value-correctness. Full per-cell results in App.~\ref{app:lammps-transfer}.}
  \label{tab:lammps-headline}
  \centering
  \small
  \begin{tabular}{llrrr}
    \toprule
    \textbf{Backbone} & \textbf{Best SIGA cell} & \textbf{Vanilla} $\uparrow$ & \textbf{Best SIGA} $\uparrow$ & \textbf{Gain} \\
    \midrule
    deepseek-v4-flash & M$+$R$+$S$+$X & $4.56$ & $\mathbf{7.78}$ & $+3.22$ \\
    Claude Sonnet 4.6 & M$+$R         & $6.33$ & $6.89$          & $+0.56$ \\
    \bottomrule
  \end{tabular}
\end{table}

\begin{rqanswer}{RQ5 (LAMMPS)}
The recipe transfers across simulator and model family, but the dominant component shifts: when scripts are already structurally complete and the bottleneck is value-correctness, procedural memory ($\mathrm{M}$, $+2.13$ on DeepSeek) and retrieval ($\mathrm{R}$, $+1.55$) carry the gain rather than the stop-hook. A DeepSeek agent at $4.56$ unguided reaches $7.78$ with the full adapter through informational scaffolding alone.
\end{rqanswer}

\section{Broader impact}
\label{sec:discussion}
\label{sec:analysis}

Lowering the configuration barrier could accelerate legitimate research (CO$_2$ storage, geothermal, induced seismicity) but could also enable less careful studies. Human expert review remains important; we frame the system as an assistant. Limitations, an extended cross-simulator analysis, the procedural-memory-tool negative result, and concrete adapter-design recommendations are collected in App.~\ref{app:discussion}.

\section{Conclusion}
\label{sec:conclusion}

We asked how far an off-the-shelf coding agent can be pushed for scientific simulator setup by wrapper-level grounding alone, taking GEOS deck authoring as our case study. On in-distribution tasks the bare harness already operates near a quality ceiling; on harder held-out compound multi-physics tasks our adapters lift mean TreeSim by roughly seven percentage points and reduce across-run variance by roughly an order of magnitude, largely by preventing the harness from occasionally returning unparseable or empty decks. Schema-driven adapters eliminate whole-block omissions, and transfer studies on OpenFOAM (30 tasks, where SIGA returns complete cases that two native agents do not) and LAMMPS (where memory and retrieval, not the completeness gate, carry the gain) show the recipe carries over while its dominant component adapts to each interface's binding constraint. As an application, SIGA brings deck authoring inside the time envelope of an experienced GEOS user, and for domain experts who are new to the simulator it compresses a multi-hour authoring task into minutes (a ${\sim}36\times$ speedup at matched deck quality on our representative task); as an initial AI-for-science contribution, it argues that the advanced tool-operating bottleneck deserves deliberate attention before agents are entrusted with harder scientific reasoning, and that adapting an existing engineered harness is a viable alternative to rebuilding agents from scratch for each new scientific target.

\medskip

{
\small
}
\bibliographystyle{abbrvnat}
\bibliography{references}

\begin{thebibliography}{36}
\providecommand{\natexlab}[1]{#1}
\providecommand{\url}[1]{\texttt{#1}}
\expandafter\ifx\csname urlstyle\endcsname\relax
  \providecommand{\doi}[1]{doi: #1}\else
  \providecommand{\doi}{doi: \begingroup \urlstyle{rm}\Url}\fi

\bibitem[Alber et~al.(2026)Alber, Chen, Sun, Isakova, Wilk, and
  Zou]{alber2025cellvoyager}
S.~Alber, B.~Chen, E.~Sun, A.~Isakova, A.~J. Wilk, and J.~Zou.
\newblock Cellvoyager: Ai compbio agent generates new insights by autonomously
  analyzing biological data.
\newblock \emph{Nature Methods}, pages 1--11, 2026.

\bibitem[Boiko et~al.(2023)Boiko, MacKnight, Kline, and
  Gomes]{boiko2023coscientist}
D.~A. Boiko, R.~MacKnight, B.~Kline, and G.~Gomes.
\newblock Autonomous chemical research with large language models.
\newblock \emph{Nature}, 624:\penalty0 570--578, 2023.
\newblock \doi{10.1038/s41586-023-06792-0}.

\bibitem[Bran et~al.(2023)Bran, Cox, Schilter, Baldassari, White, and
  Schwaller]{bran2024chemcrow}
A.~M. Bran, S.~Cox, O.~Schilter, C.~Baldassari, A.~D. White, and P.~Schwaller.
\newblock Chemcrow: Augmenting large-language models with chemistry tools,
  2023.
\newblock URL \url{https://arxiv.org/abs/2304.05376}.

\bibitem[Chen et~al.(2023)Chen, Lin, Schärli, and Zhou]{chen2024selfdebug}
X.~Chen, M.~Lin, N.~Schärli, and D.~Zhou.
\newblock Teaching large language models to self-debug, 2023.
\newblock URL \url{https://arxiv.org/abs/2304.05128}.

\bibitem[Chen et~al.(2024{\natexlab{a}})Chen, Zhu, Zhou, and
  Ren]{chen2024metaopenfoam}
Y.~Chen, X.~Zhu, H.~Zhou, and Z.~Ren.
\newblock Metaopenfoam: an llm-based multi-agent framework for cfd,
  2024{\natexlab{a}}.
\newblock URL \url{https://arxiv.org/abs/2407.21320}.

\bibitem[Chen et~al.(2024{\natexlab{b}})Chen, Chen, Ning, Zhang, Wang, Yu, Li,
  Liao, Wei, Lu, Dey, Xue, Baker, Burns, Adu-Ampratwum, Huang, Ning, Gao, Su,
  and Sun]{chen2024scienceagentbench}
Z.~Chen, S.~Chen, Y.~Ning, Q.~Zhang, B.~Wang, B.~Yu, Y.~Li, Z.~Liao, C.~Wei,
  Z.~Lu, V.~Dey, M.~Xue, F.~N. Baker, B.~Burns, D.~Adu-Ampratwum, X.~Huang,
  X.~Ning, S.~Gao, Y.~Su, and H.~Sun.
\newblock Scienceagentbench: Toward rigorous assessment of language agents for
  data-driven scientific discovery, 2024{\natexlab{b}}.
\newblock URL \url{https://arxiv.org/abs/2410.05080}.

\bibitem[{Cursor Research} et~al.(2026){Cursor Research}, Chan, Shalaby,
  Wettig, Sanger, Zhai, Ajay, Nair, Snell, Lu, Shen, Jia, Cassano, Liu, Chen,
  Wildermuth, Jackson, Li, Katz, Yao, Hejna, Warner, Vering, Frans, Danilek,
  Wright, Cen, Melas-Kyriazi, Truell, de~Jong, Jain, Schmidt, Wang,
  Muennighoff, Rybkin, Loh, Kravtsov, Yadav, Shah, Kottler, Rush, Zhang, Jain,
  Sankar, Heule, Sul, Asif, Rong, Zhu, Lin, Wu, Volkov, Zemlyanskiy, Holbrook,
  and Zhang]{cursor2026composer2}
{Cursor Research}, A.~Chan, A.~Shalaby, A.~Wettig, A.~Sanger, A.~Zhai, A.~Ajay,
  A.~Nair, C.~Snell, C.~Lu, C.~Shen, E.~Jia, F.~Cassano, H.~Liu, H.~Chen,
  H.~Wildermuth, J.~Jackson, J.~Li, J.~Katz, J.~Yao, J.~Hejna, J.~Warner,
  J.~Vering, K.~Frans, L.~Danilek, L.~Wright, L.~Cen, L.~Melas-Kyriazi,
  M.~Truell, M.~de~Jong, N.~Jain, N.~Schmidt, N.~Wang, N.~Muennighoff,
  O.~Rybkin, P.~Loh, P.~Kravtsov, R.~Yadav, S.~Shah, S.~Kottler, A.~M. Rush,
  S.~Zhang, S.~Jain, S.~Sankar, S.~Heule, S.~H. Sul, S.~Asif, V.~Rong, W.~Zhu,
  W.~Lin, Y.~Wu, Y.~Volkov, Y.~Zemlyanskiy, Z.~Holbrook, and Z.~Zhang.
\newblock Composer 2 technical report, 2026.
\newblock URL \url{https://arxiv.org/abs/2603.24477}.

\bibitem[{GEOS Development Team}(2024)]{geos2024}
{GEOS Development Team}.
\newblock {GEOS}: A multiphysics simulation framework for subsurface
  applications, 2024.
\newblock URL \url{https://github.com/GEOS-DEV/GEOS}.

\bibitem[Guilbert et~al.(2025)Guilbert, Masschelein, Goumaz, Naida, and
  Schwaller]{guilbert2025dynamate}
S.~Guilbert, C.~Masschelein, J.~Goumaz, B.~Naida, and P.~Schwaller.
\newblock Dynamate: An autonomous agent for protein-ligand molecular dynamics
  simulations, 2025.
\newblock URL \url{https://arxiv.org/abs/2512.10034}.

\bibitem[Holbrook et~al.(2026)Holbrook, Verduzco, and
  Strachan]{holbrook2026lammps}
E.~Holbrook, J.~C. Verduzco, and A.~Strachan.
\newblock Evaluating llm-generated code for domain-specific languages:
  molecular dynamics with lammps, 2026.
\newblock URL \url{https://arxiv.org/abs/2603.20630}.

\bibitem[Huang et~al.(2024)Huang, Luo, Yu, Zhang, Lei, Wei, He, Huang, Liu,
  Zhao, and Liu]{huang2024dacode}
Y.~Huang, J.~Luo, Y.~Yu, Y.~Zhang, F.~Lei, Y.~Wei, S.~He, L.~Huang, X.~Liu,
  J.~Zhao, and K.~Liu.
\newblock Da-code: Agent data science code generation benchmark for large
  language models, 2024.
\newblock URL \url{https://arxiv.org/abs/2410.07331}.

\bibitem[Lee et~al.(2026)Lee, Nair, Zhang, Lee, Khattab, and
  Finn]{lee2026metaharness}
Y.~Lee, R.~Nair, Q.~Zhang, K.~Lee, O.~Khattab, and C.~Finn.
\newblock Meta-harness: End-to-end optimization of model harnesses, 2026.
\newblock URL \url{https://arxiv.org/abs/2603.28052}.

\bibitem[Lewis et~al.(2021)Lewis, Perez, Piktus, Petroni, Karpukhin, Goyal,
  Küttler, Lewis, tau Yih, Rocktäschel, Riedel, and Kiela]{lewis2021rag}
P.~Lewis, E.~Perez, A.~Piktus, F.~Petroni, V.~Karpukhin, N.~Goyal, H.~Küttler,
  M.~Lewis, W.~tau Yih, T.~Rocktäschel, S.~Riedel, and D.~Kiela.
\newblock Retrieval-augmented generation for knowledge-intensive nlp tasks,
  2021.
\newblock URL \url{https://arxiv.org/abs/2005.11401}.

\bibitem[Li et~al.(2026)Li, Zhang, Han, Liu, Xie, Zhang, Choi, Zou, and
  Lu]{li2026agentflow}
Z.~Li, H.~Zhang, S.~Han, S.~Liu, J.~Xie, Y.~Zhang, Y.~Choi, J.~Zou, and P.~Lu.
\newblock In-the-flow agentic system optimization for effective planning and
  tool use.
\newblock In \emph{International Conference on Learning Representations
  (ICLR)}, 2026.

\bibitem[Lie et~al.(2026)Lie, Møyner, Svee, and Torben]{moyner2026jutulgpt}
K.-A. Lie, O.~Møyner, E.~Svee, and J.~Torben.
\newblock Agentic scientific simulation: Execution-grounded model construction
  and reconstruction, 2026.
\newblock URL \url{https://arxiv.org/abs/2603.00214}.

\bibitem[Lin et~al.(2026)Lin, Liu, Pan, Lin, Dou, Xi, Huang, Yan, Han, Gui, and
  Jiang]{lin2026agenticharnessengineer}
J.~Lin, S.~Liu, C.~Pan, L.~Lin, S.~Dou, Z.~Xi, X.~Huang, H.~Yan, Z.~Han,
  T.~Gui, and Y.-G. Jiang.
\newblock Agentic harness engineering: Observability-driven automatic evolution
  of coding-agent harnesses, 2026.
\newblock URL \url{https://arxiv.org/abs/2604.25850}.

\bibitem[Lu et~al.(2024)Lu, Lu, Lange, Foerster, Clune, and
  Ha]{lu2024aiscientist}
C.~Lu, C.~Lu, R.~T. Lange, J.~Foerster, J.~Clune, and D.~Ha.
\newblock The ai scientist: Towards fully automated open-ended scientific
  discovery, 2024.
\newblock URL \url{https://arxiv.org/abs/2408.06292}.

\bibitem[Narayanan et~al.(2024)Narayanan, Braza, Griffiths, Ponnapati, Bou,
  Laurent, Kabeli, Wellawatte, Cox, Rodriques, and White]{narayanan2024aviary}
S.~Narayanan, J.~D. Braza, R.-R. Griffiths, M.~Ponnapati, A.~Bou, J.~Laurent,
  O.~Kabeli, G.~Wellawatte, S.~Cox, S.~G. Rodriques, and A.~D. White.
\newblock Aviary: training language agents on challenging scientific tasks,
  2024.
\newblock URL \url{https://arxiv.org/abs/2412.21154}.

\bibitem[Ni and Buehler(2023)]{ni2024mechagents}
B.~Ni and M.~J. Buehler.
\newblock Mechagents: Large language model multi-agent collaborations can solve
  mechanics problems, generate new data, and integrate knowledge, 2023.
\newblock URL \url{https://arxiv.org/abs/2311.08166}.

\bibitem[Ning et~al.(2026)Ning, Tieu, Fu, Wei, Li, Bei, Zou, Ai, Liu, Li, Chen,
  Zhao, Yang, Li, Qian, Li, Lin, Zeng, Qiu, Chen, Sun, Yang, Wang, Pan, Yang,
  Zhang, Fang, Cui, Cao, Chen, Sun, Chen, Srinivasan, Mathur, Xia, Li, Yan, Lu,
  Zhang, Zhang, Tong, and He]{ning2026codeagentharness}
X.~Ning, K.~Tieu, D.~Fu, T.~Wei, Z.~Li, Y.~Bei, J.~Zou, M.~Ai, Z.~Liu, T.-W.
  Li, L.~Chen, Y.~Zhao, K.~Yang, B.~Li, C.~Qian, G.~Li, X.~Lin, Z.~Zeng,
  R.~Qiu, S.~Chen, Y.~Sun, X.~Yang, R.~Wang, R.~Pan, C.~Yang, D.~Zhang,
  L.~Fang, Z.~Cui, Y.~Cao, P.~Chen, D.~Sun, R.~Chen, M.~Srinivasan, N.~Mathur,
  Y.~Xia, H.~Li, H.~Yan, P.~Lu, L.~Zhang, T.~Zhang, H.~Tong, and J.~He.
\newblock Code as agent harness, 2026.
\newblock URL \url{https://arxiv.org/abs/2605.18747}.

\bibitem[Pandey et~al.(2025)Pandey, Xu, Wang, and Chu]{pandey2025openfoamgpt}
S.~Pandey, R.~Xu, W.~Wang, and X.~Chu.
\newblock Openfoamgpt: A retrieval-augmented large language model (llm) agent
  for openfoam-based computational fluid dynamics.
\newblock \emph{Physics of Fluids}, 37\penalty0 (3), Mar. 2025.
\newblock ISSN 1089-7666.
\newblock \doi{10.1063/5.0257555}.
\newblock URL \url{http://dx.doi.org/10.1063/5.0257555}.

\bibitem[Park et~al.(2026)Park, Moon, and Ryu]{mcpsim2026}
D.~Park, H.~Moon, and S.~Ryu.
\newblock A self-correcting multi-agent {LLM} framework for language-based
  physics simulation and explanation.
\newblock \emph{npj Artificial Intelligence}, 2\penalty0 (1):\penalty0 10,
  2026.
\newblock \doi{10.1038/s44387-025-00057-z}.

\bibitem[Shi et~al.(2026)Shi, A, Shao, Huang, An, Xin, Shen, Wang, Na, Huang,
  and Jing]{shi2026mdagent2}
Z.~Shi, H.~A, Y.~Shao, D.~Huang, H.~An, C.~Xin, H.~Shen, Z.~Wang, Y.~Na,
  G.~Huang, and X.~Jing.
\newblock Mdagent2: Large language model for code generation and knowledge q\&a
  in molecular dynamics, 2026.
\newblock URL \url{https://arxiv.org/abs/2601.02075}.

\bibitem[Tang et~al.(2025)Tang, Xu, Wang, Guo, Shao, Chen, Zhang, Wang, Zhang,
  Wan, Zhang, Bai, Yin, Torr, Wang, and Jin]{tang2026eigenagent}
X.~Tang, W.~Xu, Y.~Wang, Z.~Guo, D.~Shao, J.~Chen, C.~Zhang, Z.~Wang, L.~Zhang,
  G.~Wan, W.~Zhang, L.~Bai, Z.~Yin, P.~Torr, H.~Wang, and D.~Jin.
\newblock Eigen-1: Adaptive multi-agent refinement with monitor-based rag for
  scientific reasoning, 2025.
\newblock URL \url{https://arxiv.org/abs/2509.21193}.

\bibitem[Thompson et~al.(2022)Thompson, Aktulga, Berger, Bolintineanu, Brown,
  Crozier, in~'t Veld, Kohlmeyer, Moore, Nguyen, Shan, Stevens, Tranchida,
  Trott, and Plimpton]{LAMMPS}
A.~P. Thompson, H.~M. Aktulga, R.~Berger, D.~S. Bolintineanu, W.~M. Brown,
  P.~S. Crozier, P.~J. in~'t Veld, A.~Kohlmeyer, S.~G. Moore, T.~D. Nguyen,
  R.~Shan, M.~J. Stevens, J.~Tranchida, C.~Trott, and S.~J. Plimpton.
\newblock {LAMMPS} - a flexible simulation tool for particle-based materials
  modeling at the atomic, meso, and continuum scales.
\newblock \emph{Comp. Phys. Comm.}, 271:\penalty0 108171, 2022.
\newblock \doi{10.1016/j.cpc.2021.108171}.

\bibitem[Wang et~al.(2025)Wang, Li, Song, Xu, Tang, Zhuge, Pan, Song, Li,
  Singh, Tran, Li, Ma, Zheng, Qian, Shao, Muennighoff, Zhang, Hui, Lin,
  Brennan, Peng, Ji, and Neubig]{wang2024openhands}
X.~Wang, B.~Li, Y.~Song, F.~F. Xu, X.~Tang, M.~Zhuge, J.~Pan, Y.~Song, B.~Li,
  J.~Singh, H.~H. Tran, F.~Li, R.~Ma, M.~Zheng, B.~Qian, Y.~Shao,
  N.~Muennighoff, Y.~Zhang, B.~Hui, J.~Lin, R.~Brennan, H.~Peng, H.~Ji, and
  G.~Neubig.
\newblock Openhands: An open platform for ai software developers as generalist
  agents, 2025.
\newblock URL \url{https://arxiv.org/abs/2407.16741}.

\bibitem[Weller et~al.(1998)Weller, Tabor, Jasak, and
  Fureby]{weller1998tensorial}
H.~G. Weller, G.~Tabor, H.~Jasak, and C.~Fureby.
\newblock A tensorial approach to computational continuum mechanics using
  object-oriented techniques.
\newblock \emph{Computers in physics}, 12\penalty0 (6):\penalty0 620--631,
  1998.

\bibitem[Yamada et~al.(2025)Yamada, Lange, Lu, Hu, Lu, Foerster, Clune, and
  Ha]{yamada2025ai}
Y.~Yamada, R.~T. Lange, C.~Lu, S.~Hu, C.~Lu, J.~Foerster, J.~Clune, and D.~Ha.
\newblock The ai scientist-v2: Workshop-level automated scientific discovery
  via agentic tree search, 2025.
\newblock URL \url{https://arxiv.org/abs/2504.08066}.

\bibitem[Yang et~al.(2024{\natexlab{a}})Yang, Jimenez, Wettig, Lieret, Yao,
  Narasimhan, and Press]{yang2024sweagent}
J.~Yang, C.~E. Jimenez, A.~Wettig, K.~Lieret, S.~Yao, K.~Narasimhan, and
  O.~Press.
\newblock Swe-agent: Agent-computer interfaces enable automated software
  engineering, 2024{\natexlab{a}}.
\newblock URL \url{https://arxiv.org/abs/2405.15793}.

\bibitem[Yang et~al.(2024{\natexlab{b}})Yang, Yu, Zhang, Cao, Xu, Zhang,
  Gonzalez, and Cui]{yang2024bot}
L.~Yang, Z.~Yu, T.~Zhang, S.~Cao, M.~Xu, W.~Zhang, J.~E. Gonzalez, and B.~Cui.
\newblock Buffer of thoughts: Thought-augmented reasoning with large language
  models, 2024{\natexlab{b}}.
\newblock URL \url{https://arxiv.org/abs/2406.04271}.

\bibitem[Yang et~al.(2026)Yang, Gong, Huang, Yang, Zhou, Huang, Li, Gao, Dai,
  Liu, Qiu, Yang, Chen, Yang, and Luo]{yang2026skillopt}
Y.~Yang, Z.~Gong, W.~Huang, Q.~Yang, Z.~Zhou, Z.~Huang, Y.~Li, X.~Gao, Q.~Dai,
  B.~Liu, K.~Qiu, Y.~Yang, D.~Chen, X.~Yang, and C.~Luo.
\newblock Skillopt: Executive strategy for self-evolving agent skills, 2026.
\newblock URL \url{https://arxiv.org/abs/2605.23904}.

\bibitem[Yue et~al.(2025{\natexlab{a}})Yue, Somasekharan, Zhang, Cao, Chen, Di,
  and Pan]{yue2025foamagent}
L.~Yue, N.~Somasekharan, T.~Zhang, Y.~Cao, Z.~Chen, S.~Di, and S.~Pan.
\newblock Foam-agent: Towards automated intelligent cfd workflows,
  2025{\natexlab{a}}.
\newblock URL \url{https://arxiv.org/abs/2505.04997}.

\bibitem[Yue et~al.(2025{\natexlab{b}})Yue, Somasekharan, Zhang, Cao, and
  Pan]{foamagent2025}
L.~Yue, N.~Somasekharan, T.~Zhang, Y.~Cao, and S.~Pan.
\newblock Foam-agent 2.0: An end-to-end composable multi-agent framework for
  automating cfd simulation in openfoam, 2025{\natexlab{b}}.
\newblock URL \url{https://arxiv.org/abs/2509.18178}.

\bibitem[Zhang and Sun(2026)]{zhang2026scinav}
T.~Zhang and H.~Sun.
\newblock Scinav: A general agent framework for scientific coding tasks, 2026.
\newblock URL \url{https://arxiv.org/abs/2603.20256}.

\bibitem[Zhang et~al.(2025)Zhang, Liu, Xin, and Jiao]{zhan2025mooseagent}
T.~Zhang, Z.~Liu, Y.~Xin, and Y.~Jiao.
\newblock Mooseagent: A llm based multi-agent framework for automating moose
  simulation, 2025.
\newblock URL \url{https://arxiv.org/abs/2504.08621}.

\bibitem[Zhao et~al.(2026)Zhao, Chandrasekhar, and
  Farimani]{zhao2026polyjarvis}
A.~Zhao, A.~Chandrasekhar, and A.~B. Farimani.
\newblock Polyjarvis: Llm agent for autonomous polymer md simulations, 2026.
\newblock URL \url{https://arxiv.org/abs/2604.02537}.

\end{thebibliography}


\appendix

\section{Benchmark details}
\label{app:bench}

\subsection{Task pool, splits, and hygiene}
The 46-task pool is mined from GEOS advanced examples and tutorial decks. Ten tasks are held out as the held-out-evaluation pool (the same ten reported as held-out-eval throughout; never used as in-context demonstrations during the main GEOS runs): \texttt{AdvancedExampleCasedThermoElasticWellbore}, \texttt{AdvancedExamplePureThermalDiffusionWellbore}, \texttt{AdvancedExampleThermoPoroElasticWellbore}, \texttt{AdvancedExampleViscoExtendedDruckerPrager}, \texttt{ExampleIsothermalHystInjection}, \texttt{ExampleMCCWellbore}, \texttt{ExampleProppantTest}, \texttt{ExamplesingleFracCompression}, \texttt{ExampleVerticalPoroElastoPlasticWellbore}, \texttt{TutorialHydraulicFractureWithAdvancedXML}. The remaining 36 are split by odd/even index (sorted by training-run TreeSim, failures-as-zero) into an 18-task distillation corpus and a 17-task validation-selection set (one task dropped). Ground-truth decks live at \texttt{/data/shared/geophysics\_agent\_data/data/eval/experiments\_gt/}.

\subsection{Spec versions (v1 vs v2)}
Two instruction sets exist: v1 (original mined specs, used in pre-2026-04-21 deepseek runs) and v2 (cleaner, more self-contained rewrites, used in all post-D-004 runs). Mid-project comparisons that crossed the versions were contaminated; re-runs on v2 re-anchored the canonical baselines. All numbers in this paper use v2.

\section{Bottleneck-analysis pipeline (full)}
\label{app:bottleneck}

\textbf{Stage 1 (LLM-free extractor).} For each (cell, run, task), the extractor walks the recursive \texttt{treesim\_detail} tree from the scorer and ranks subtrees by impact $= (1 - \mathrm{score}) \cdot (n_{\mathrm{gt\_children}} + 1)$, returning the top-$K$. It also extracts trajectory features from \texttt{events.jsonl}: tool counts, file re-reads, xmllint MCP calls, edit churn, grep/glob query distributions.

\textbf{Stage 2 (\texttt{deepseek-v4-flash} per-task classifier).} The extractor's output, plus a focused GT-vs-generated XML excerpt for the worst-scoring section when its score is below 0.7, is sent to \texttt{deepseek-v4-flash} with a strict-JSON output schema. The model returns: \texttt{failure\_category} $\in \{$\texttt{missing\_block}, \texttt{extra\_block}, \texttt{hallucinated\_extras}, \texttt{structural\_mismatch}, \texttt{bad\_attribute\_value}, \texttt{partial\_implementation}, \texttt{wrong\_constitutive}, \texttt{no\_failure}$\}$, a \texttt{primary\_failure\_section}, a one-sentence \texttt{root\_cause}, a \texttt{trajectory\_evidence} citation, and a \texttt{would\_have\_helped} hypothesis.

\textbf{Stage 3 (\texttt{deepseek-v4-pro} synthesis).} Per-cell category and section distributions, section ``failure weight'' $= \sum_{\mathrm{tasks}}(1 - \mathrm{TreeSim})$ grouped by primary failing section, and per-task baseline-vs-best deltas are sent to \texttt{deepseek-v4-pro} with an instruction to write a paper-grade synthesis anchored on a chosen baseline/best pair. Total cost across all evaluations was ${\sim}650$ flash calls plus 4 pro narratives, ${\sim}\$5\text{--}7$ in DeepSeek API spend.

Code: \texttt{scripts/bottleneck/\{extract,llm\_per\_task,aggregate\}.py}.

\subsection{Bottleneck failure-category counts (full)}
\label{app:bottleneck-counts}

\begin{table}[h]
  \caption{Failure-category counts per cell; categories with all-zero or unreported counts (\texttt{no\_failure}, \texttt{wrong\_constitutive}) are omitted. val panel uses $n=51$ tasks per cell ($3{\times}17$); held-out-eval panel uses $n=29$--$30$ per cell ($3{\times}10$ minus a small number of LLM-judge parse-failures).}
  \label{tab:bottleneck}
  \centering
  \footnotesize
  \resizebox{0.98\textwidth}{!}{%
  \begin{tabular}{l|ccccc|cccccc}
    \toprule
    & \multicolumn{5}{c|}{\textbf{val ($n=51$)}} & \multicolumn{6}{c}{\textbf{held-out-eval ($n=30$)}} \\
    \textbf{Category} & Vanilla & S+M & X+M & S+X & SE & Vanilla & X+M & S+X & S+X+M & SE-prose & SE \\
    \midrule
    bad\_attribute\_value     & 12 & 12 & 11 & 15 & 9  & 5 & 7 & 5 & 8 & 0 & 0 \\
    extra\_block              &  9 &  8 & 11 &  5 & 10 & 9 & 4 & 4 & 6 & 7 & 5 \\
    hallucinated\_extras      &  4 &  8 &  7 &  - &  - & 0 & 0 & 0 & 0 & 3 & 4 \\
    missing\_block            &  6 &  4 &  3 &  2 &  - & 4 & 5 & 7 & 7 & 7 & 4 \\
    partial\_implementation   &  7 &  9 &  6 & 10 &  9 & 1 & 0 & 4 & 3 & 3 & 5 \\
    structural\_mismatch      &  6 &  8 &  7 & 11 &  - & 8 & 6 & 6 & 5 & 3 & 5 \\
    \bottomrule
  \end{tabular}}
\end{table}


\begin{table}[h]
  \caption{Per-task held-out-eval TreeSim (mean across 3 runs), sorted ascending by Vanilla score. The aggregate $+0.069$ Vanilla$\to$SE gain is concentrated in the top two non-universal-failure tasks (rows 2--3); the remaining seven tasks show within-noise differences across cells.}
  \label{tab:per-task-icl10}
  \centering
  \footnotesize
  \resizebox{0.98\textwidth}{!}{%
  \begin{tabular}{lrrrrrr}
    \toprule
    \textbf{Task} & \textbf{Vanilla} & \textbf{X+M} & \textbf{S+X} & \textbf{S+X+M} & \textbf{SE-prose} & \textbf{SE} \\
    \midrule
    \texttt{TutorialHydraulicFractureWithAdvancedXML}      & 0.013 & 0.013 & 0.013 & 0.013 & 0.013 & 0.013 \\
    \texttt{AdvancedExampleThermoPoroElasticWellbore}      & 0.355 & 0.680 & 0.681 & 0.708 & 0.743 & \textbf{0.761} \\
    \texttt{ExampleProppantTest}                           & 0.541 & 0.810 & 0.809 & 0.799 & 0.817 & \textbf{0.825} \\
    \texttt{ExampleIsothermalHystInjection}                & 0.755 & 0.747 & 0.751 & 0.750 & 0.769 & 0.717 \\
    \texttt{AdvancedExampleCasedThermoElasticWellbore}     & 0.847 & 0.807 & 0.923 & 0.919 & 0.877 & 0.886 \\
    \texttt{ExamplesingleFracCompression}                  & 0.891 & 0.887 & 0.904 & 0.931 & 0.929 & 0.928 \\
    \texttt{ExampleVerticalPoroElastoPlasticWellbore}      & 0.909 & 0.903 & 0.906 & 0.891 & 0.834 & 0.944 \\
    \texttt{ExampleMCCWellbore}                            & 0.935 & 0.924 & 0.908 & 0.905 & 0.905 & 0.941 \\
    \texttt{AdvancedExamplePureThermalDiffusionWellbore}   & 0.963 & 0.922 & 0.956 & 0.947 & 0.864 & 0.880 \\
    \texttt{AdvancedExampleViscoExtendedDruckerPrager}     & 0.986 & 0.991 & 0.963 & 0.964 & 0.999 & 0.996 \\
    \bottomrule
  \end{tabular}}
\end{table}

\section{Cross-model and cross-harness: full panels}
\label{app:cross-model-detail}
\label{app:cross-model}

\textbf{Setup.} Cross-model: $n=1$ run of Vanilla, X+M, SE on val with two backbones (\texttt{minimax-m2.7}, \texttt{google/gemini-3-flash-preview}) via OpenRouter, harness held at Claude Code, default-off harness env (\texttt{GEOS\_HOOK\_POSTTOOLUSE} unset, autocamp-experiment-state parity). Cross-harness: $n=3$ runs of Vanilla and X+M on val with \texttt{deepseek-v4-flash}, harness varied between Claude Code and OpenHands. Headline DSv4 numbers from Table~\ref{tab:main-results} included for comparison.

\begin{table}[h]
  \caption{Full cross-model and cross-harness panel. Counts in parentheses are task failures out of 17.}
  \label{tab:cross-cutting-full}
  \centering
  \small
  \resizebox{0.98\textwidth}{!}{%
  \begin{tabular}{llccc}
    \toprule
    \textbf{Harness} & \textbf{Backbone} & \textbf{Vanilla} & \textbf{X+M} & \textbf{SE} \\
    \midrule
    CC & \texttt{deepseek-v4-flash} ($n=3$) & $0.910 \pm 0.024$ & $0.921 \pm 0.007$ & $0.919 \pm 0.020$ \\
    CC & \texttt{minimax-m2.7} ($n=1$)        & $0.821$ ($1$ fail) & $0.867$ ($0$ fails) & $0.861$ ($0$ fails) \\
    CC & \texttt{gemini-3-flash-preview} ($n=1$) & $0.768$ ($0$ fails) & $0.797$ ($0$ fails) & $0.757$ ($1$ fail) \\
    OH & \texttt{deepseek-v4-flash} ($n=3$) & $0.856 \pm 0.061$ & $0.881 \pm 0.023$ & -- \\
    \bottomrule
  \end{tabular}}
\end{table}

\textbf{R-factor cross-model corroboration} (older runs). On a 35-task superset that overlaps val, the retrieval plugin lifts $+0.175$ on \texttt{deepseek-v3.2} (paired 35 tasks) and $+0.115$ on \texttt{minimax-m2.7} (paired 15 tasks). Retrieval thus helps on weaker backbones even though its val main effect on the near-ceiling \texttt{deepseek-v4-flash} is slightly negative; we report these older runs only as cross-model corroboration and defer to the headline val results above.

\textbf{Harness-less floor.} A one-shot direct-prompting baseline on \texttt{minimax-m2.7} (\texttt{scripts/harnessless\_eval.py}; \texttt{deepseek-v4-flash} not yet run in this configuration) reaches TreeSim $= 0.333$ on val. We treat this as a model-class lower bound rather than a per-model floor.

\textbf{Memory-as-retrieval negative result.} An earlier version of our system exposed primer-like content via an MCP \texttt{memory\_lookup} tool with embedding top-$k$ retrieval over the distilled corpus. Across every test-set run in which this tool was available (the A4, A4$'$, A5, M3-g conditions of the prior campaign), the agent called the tool zero times. The tool was verified functional. Delivering the same content through the always-on \texttt{--append-system-prompt} channel (i.e., the M factor in this paper) is what produced any lift. We retain this as a clean negative result on retrieval-based memory for this task--model class.

\section{OpenFOAM transfer study}
\label{app:openfoam-transfer}

\paragraph{Motivation.}
The main paper studies SIGA on GEOS, where the executable interface is a structured XML deck with an explicit XSD schema. To test whether the same adapter recipe transfers beyond GEOS and beyond XML-backed simulator interfaces, we ran an OpenFOAM case-authoring companion study in a separate repository path (\texttt{repo3\_openfoam}). The headline leaderboard is Table~\ref{tab:openfoam-main} in the main text; this appendix gives the adaptation, the benchmark and metric, the two native baselines, the factor readout, and the cost accounting.

\paragraph{OpenFOAM adaptation of \texorpdfstring{$R/S/X/M$}{R/S/X/M}.}
We ported the same four binary factors to OpenFOAM case authoring. \textbf{R} replaced the GEOS RAG collections with three ChromaDB collections over OpenFOAM tutorial structure, detailed case snippets, and command/help text. \textbf{M} replaced the GEOS memory cheatsheet with a lightweight always-on primer describing the standard OpenFOAM case skeleton (\texttt{0/}, \texttt{constant/}, \texttt{system/}) and the relevant tutorial/source-tree locations. \textbf{S} replaced the GEOS XML stop-hook with an OpenFOAM stop-hook that blocks turn completion if required files are missing or if generated dictionaries fail heuristic structural checks. \textbf{X} replaced the \texttt{xmllint} MCP with an agent-callable OpenFOAM validator using the same case checks as the hook. Because OpenFOAM lacks a canonical XSD-style schema for the benchmark dictionaries, the OpenFOAM validator checks required-file presence, balanced delimiters, \texttt{FoamFile} headers, and key dictionary sections rather than schema compliance.

\paragraph{Benchmark and metric.}
The benchmark (\texttt{foamgpt\_subset\_seed42\_n30\_hybrid}) contains 30 tasks sampled from a FoamGPT-derived pool, spanning incompressible, compressible, multiphase, combustion, heat-transfer, lagrangian, mesh, DNS, molecular-dynamics, and discrete-method solvers. Each task specifies a set of required files. We score a run with a file-text-and-coverage metric: each missing required file receives score 0; present files are compared to ground truth via normalized text similarity; and the case score is
\[
\mathrm{Score} = 0.7 \cdot \mathrm{mean\_similarity} + 0.3 \cdot \mathrm{coverage}.
\]
This is analogous to failures-as-zero reporting but is not TreeSim. All three systems were re-run on the same date on \texttt{deepseek/deepseek-v4-flash} via OpenRouter with robust token and tool-call instrumentation added beforehand; cost is recomputed per row from logged input/output tokens at OpenRouter DeepSeek V4 Flash effective pricing (input \$0.0983/M, output \$0.1966/M). Headline scores shifted only modestly from a pre-instrumentation run (e.g., repo3 R+S $0.887\to0.870$), within run-to-run variance of a stochastic model; the instrumentation change affects only token/cost accounting, not scoring.

\paragraph{Native baselines.}
We compare against two from-scratch OpenFOAM agents, Foam-Agent~\citep{foamagent2025} and MetaOpenFOAM~\citep{chen2024metaopenfoam}, both in lint-only execution mode. A caveat that matters for interpretation: neither native workflow is lint-only by design (Foam-Agent runs a fuller plan--write--execute--review loop); the stable comparison available in our environment used \texttt{execution\_mode=lint\_only}, since execute-mode runs failed to yield usable benchmark outputs. We therefore treat both as constrained baselines rather than full head-to-heads against their intended execution-coupled workflows. Token counts for both are sourced from real provider usage metadata, aggregated across every LLM service instance per task; tool-call columns are not directly comparable across systems (different definitions) and we do not compare on them.

\paragraph{Coverage and failure analysis.}
The reliability contrast is in coverage. Every SIGA cell, including Vanilla, holds 30/30 full required-file coverage with 0 zero-score tasks. Foam-Agent reaches 19/30 full coverage with 8 zero-score tasks; several of its zero scores are completed runs that omitted required paths (e.g.\ \texttt{freeSpacePeriodic}, \texttt{squareBump}, \texttt{supersonicCorner}, \texttt{aachenBomb}) rather than crashes, which the coverage metric penalizes heavily. MetaOpenFOAM reaches 10/30 full coverage with 12 zero-score tasks, its weakness again exact required-file coverage. In other words, the SIGA harness eliminates the missing-required-file failure mode that dominates both native agents' score collapses, which is the OpenFOAM analogue of the GEOS hard-tail rescue.

\paragraph{Factor-style readout.}
Using the eight SIGA factorial cells and reporting single-run descriptive main effects on mean score:
\[
R: +0.005,\quad
S: +0.168,\quad
X: +0.007,\quad
M: -0.007.
\]
$\mathrm{S}$ is the dominant effect by an order of magnitude over the others, consistent with the GEOS finding that a hard end-of-turn completeness gate, rather than optional retrieval or optional validation, is the portable intervention. The $\mathrm{S}$ factor appears in all five top-scoring cells (R+S, S+X, R+S+X+M, S+X+M, S+M).

\paragraph{Cost.}
The SIGA harness is the most expensive of the three per task: total \$23.20 across all nine cells (270 task-runs), because its agentic loop resends the growing context each tool turn and the OpenRouter DeepSeek route reports no prompt-cache reads. Foam-Agent (\$0.30 total) and MetaOpenFOAM (\$0.17 total) are roughly an order of magnitude cheaper per task, trading cost for the coverage and quality the SIGA harness buys with more context. Per-row wall-clock and cost are in Table~\ref{tab:openfoam-main}.

\paragraph{Interpretation.}
The OpenFOAM result is single-run and scored with a file-text-and-coverage metric rather than TreeSim, so we read it as transfer evidence rather than a second full benchmark. What it shows, now at 30 tasks and against two native agents, is that the SIGA recipe is not tied to XML or to GEOS-specific assets: the transferable piece is the workflow logic, and the dominant, most portable intervention is forced end-of-turn verification that prevents silent incompleteness.

\section{LAMMPS transfer study}
\label{app:lammps-transfer}

\paragraph{Motivation.}
The OpenFOAM study (App.~\ref{app:openfoam-transfer}) showed the recipe extends to a schema-free, text-file simulator. To probe whether it also transfers across model family and into a physically richer domain, we ran a LAMMPS molecular-dynamics input-script authoring study with two backbone models across 12 factor configurations. Both backbones run inside the Claude Code harness; ``Claude'' denotes the Claude Sonnet 4.6 backbone model (also used as the LLM judge), not the harness, and ``DeepSeek'' denotes \texttt{deepseek-v4-flash}. The Claude Code harness is used in every experiment in this paper except the OpenHands cross-harness run (App.~\ref{app:cross-model-detail}). LAMMPS differs from GEOS and OpenFOAM in several ways: its input is a sequential command script with strict ordering rules and no formal schema; a large curated example library ships with the source; and the physics space is broader (thermostats, barostats, 2D simulations, NEMD, electrostatics, mechanical deformation). We treat the result as transfer evidence rather than a second benchmark.

\paragraph{LAMMPS adaptation of \texorpdfstring{$R/S/X/M$}{R/S/X/M}.}
We ported the same four binary factors to LAMMPS input authoring. \textbf{R} replaced the GEOS RAG collections with three ChromaDB collections over LAMMPS example scripts, documentation RST files, and command syntax extracted from the LAMMPS source tree. \textbf{M} replaced the GEOS XML cheatsheet with a LAMMPS-specific primer covering required command ordering, unit conventions, and ten task-specific critical pitfalls (e.g.\ LJ lattice density semantics, region-before-\texttt{create\_box} ordering, unfix before switching integrators, the SLLOD Couette pattern, MSD compute syntax). \textbf{S} replaced the GEOS XML stop-hook with a LAMMPS structural stop-hook that blocks turn completion if the generated script fails heuristic structural checks. \textbf{X} replaced \texttt{xmllint} with an agent-callable LAMMPS script validator applying the same checks as the hook. Because LAMMPS has no canonical schema for these tasks, the validator checks command presence, correct units and \texttt{atom\_style}, ensemble type, task-specific physics commands (\texttt{fix indent}, \texttt{fix deform}, \texttt{compute msd}, \texttt{fix nvt/sllod}, \texttt{fix npt}), temperature ranges, and minimum run lengths.

\paragraph{Benchmark and metric.}
The benchmark contains 9 tasks spanning a range of MD physics: Lennard-Jones melt from an FCC lattice (\texttt{lj\_melt}, \texttt{lj\_melt\_minimal}), LJ FCC solid NVT equilibration (\texttt{lj\_solid}), 2D crack propagation (\texttt{crack\_2d}), 2D nanoindentation (\texttt{lj\_indent}), Couette flow via SLLOD (\texttt{couette\_flow}), mean-square-displacement diffusion (\texttt{msd\_diffusion}), TIP3P water NVT with long-range electrostatics (\texttt{nvt\_water}), and uniaxial tension with lateral NPT relaxation (\texttt{uniaxial\_tension}). Each task provides a detailed natural-language specification; the agent writes the input script without executing LAMMPS. We use a two-stage metric. Stage 1 (structural) is a deterministic regex-based check over 11--15 per-task criteria (required commands, expected command patterns, task-specific physics markers, coarse temperature/step-count ranges); it verifies presence and rough plausibility, not value-correctness. Stage 2 (LLM judge) gives the specification, a ground-truth script, and the agent script to Claude Sonnet 4.6 and elicits an integer score $0$--$10$ with rationale. LLM-judge scores are not fully deterministic at temperature 0 (per-call variance through OpenRouter); we report a single evaluation pass and treat the scores as approximate rankings. Mean LLM-judge score is the primary metric; structural score is a secondary reliability indicator (near-ceiling, $\geq 0.976$, for all 12 cells).

\begin{table}[t]
  \caption{Per-cell LAMMPS LLM-judge scores ($0$--$10$), 9 tasks, single run, two backbones. Structural scores are near-ceiling ($\geq 0.976$) for all cells; $\dagger$ marks the two cells with structural $< 1.0$. Best mean per backbone in bold.}
  \label{tab:lammps-percell}
  \centering
  \footnotesize
  \resizebox{0.98\textwidth}{!}{%
  \begin{tabular}{llrrrrrrrrrr}
    \toprule
    \textbf{Backbone} & \textbf{Config} & \textbf{couette} & \textbf{crack\_2d} & \textbf{lj\_indent} & \textbf{lj\_melt} & \textbf{lj\_melt\_min} & \textbf{lj\_solid} & \textbf{msd\_diff} & \textbf{nvt\_water} & \textbf{tension} & \textbf{mean} \\
    \midrule
    Claude Sonnet 4.6   & Vanilla            & 7 & 3 & 3 & 7 & 9 & 9 & 6 & 5 & 8 & $6.33$ \\
    Claude Sonnet 4.6   & M$+$R              & 6 & 6 & 2 & 7 & 9 & 9 & 9 & 8 & 6 & $\mathbf{6.89}$ \\
    Claude Sonnet 4.6   & M$+$S              & 6 & 6 & 2 & 5 & 9 & 8 & 8 & 7 & 6 & $6.33$ \\
    Claude Sonnet 4.6   & M$+$R$+$S          & 7 & 1 & 4 & 6 & 8 & 7 & 8 & 7 & 8 & $6.22$ \\
    Claude Sonnet 4.6   & M$+$S$+$X $\dagger$ & 7 & 2 & 3 & 2 & 6 & 8 & 9 & 7 & 8 & $5.78$ \\
    Claude Sonnet 4.6   & M$+$R$+$S$+$X      & 8 & 3 & 4 & 8 & 7 & 9 & 9 & 5 & 9 & $\mathbf{6.89}$ \\
    \midrule
    DeepSeek v4-flash & Vanilla $\dagger$  & 6 & 2 & 2 & 2 & 9 & 8 & 3 & 3 & 6 & $4.56$ \\
    DeepSeek v4-flash & M$+$R              & 7 & 5 & 6 & 9 & 4 & 8 & 7 & 8 & 8 & $6.89$ \\
    DeepSeek v4-flash & M$+$S              & 5 & 2 & 5 & 8 & 6 & 8 & 8 & 6 & 7 & $6.11$ \\
    DeepSeek v4-flash & M$+$R$+$S          & 8 & 4 & 3 & 8 & 9 & 9 & 5 & 8 & 6 & $6.67$ \\
    DeepSeek v4-flash & M$+$S$+$X          & 6 & 2 & 2 & 9 & 9 & 8 & 5 & 6 & 7 & $6.00$ \\
    DeepSeek v4-flash & M$+$R$+$S$+$X      & 8 & 3 & 8 & 9 & 9 & 9 & 8 & 7 & 9 & $\mathbf{7.78}$ \\
    \bottomrule
  \end{tabular}}
\end{table}

\paragraph{Factor-style readout.}
Using the 6-cell subset per backbone and reporting single-run descriptive main effects:
\begin{align*}
\text{Claude Sonnet 4.6:}\quad & R{:}\,{+}0.52,\ \ S{:}\,{-}0.30,\ \ X{:}\,{-}0.10,\ \ M{:}\,{+}0.09;\\
\text{DeepSeek v4-flash:}\quad & R{:}\,{+}1.55,\ \ S{:}\,{+}0.91,\ \ X{:}\,{+}0.83,\ \ M{:}\,{+}2.13.
\end{align*}
On DeepSeek, $\mathrm{M}$ exhibits the largest descriptive main effect: the primer contributes an average marginal gain of over two points relative to the unguided baseline, and $\mathrm{R}$ is second, indicating that access to the LAMMPS example and documentation corpus matters substantially for this model. On Claude, the baseline is already strong and all effects are small; the only reliably positive factor is $\mathrm{R}$. This contrasts with GEOS and OpenFOAM, where $\mathrm{S}$ was dominant.

\paragraph{Why the adapters help, and why structural scores are near-ceiling.}
Both backbones reach near-identical structural scores ($\geq 0.976$) across all configurations: both reliably include all required commands and parameter fields regardless of which adapters are active. The LLM judge, which compares the actual values assigned to those parameters against ground truth, is where they diverge. Failure in this domain is therefore never whole-section omission; it is the correctness of specific values, physical constants, lattice densities, ensemble parameters, command-argument ordering. The failure-mode analysis across all 12 agents points to two correctable classes. $\mathrm{M}$ (the primer) addresses knowledge errors: LJ unit-convention mistakes (passing a lattice constant as a reduced density, real argon parameters in an LJ-unit script) and command-ordering errors (\texttt{create\_box} before defining a region, two integrators active because NVE was not unfixed before NVT). $\mathrm{R}$ (retrieval) addresses syntax-precision errors: invalid lattice style names, incorrect \texttt{fix} argument ordering, undefined \texttt{compute} references, all exact-token mistakes a retrieved working example corrects immediately.

\paragraph{Why the DeepSeek gain is larger.}
The $3.2$-point DeepSeek gain ($4.56\to7.78$) versus the $0.56$-point Claude gain ($6.33\to6.89$) reflects a difference in baseline knowledge. Claude vanilla already assigns mostly correct values; its residual errors concentrate on tasks needing precise geometric reasoning (\texttt{crack\_2d} notch placement), which neither $\mathrm{M}$ nor $\mathrm{R}$ fully resolves. DeepSeek vanilla makes more of the knowledge-correctable unit/ordering/syntax errors that $\mathrm{M}$ and $\mathrm{R}$ directly target, so the adapters narrow the baseline gap and let DeepSeek match or exceed guided Claude. Per-task, the gains concentrate where the DeepSeek baseline is lowest (\texttt{lj\_melt}, \texttt{crack\_2d}, \texttt{lj\_indent}, \texttt{msd\_diffusion}, \texttt{nvt\_water}); the highest-baseline tasks (\texttt{lj\_solid}, \texttt{lj\_melt\_minimal}) are essentially unchanged or mildly hurt, since they closely match common MD examples. \texttt{crack\_2d} remains the hardest task across both backbones and every adapter combination: getting the notch-coordinate logic right is a geometric-reasoning problem, not a knowledge lookup.

\paragraph{Interpretation.}
The LAMMPS result replicates the core finding, that the plugin recipe transfers, but shifts the dominant mechanism. In GEOS and OpenFOAM, $\mathrm{S}$ (forced end-of-turn verification) was the largest reliability effect because it prevented silent failure to produce required outputs. In LAMMPS we observed little silent failure; performance differences were driven by the correctness of assigned parameter values, so $\mathrm{M}$ and $\mathrm{R}$ carry the quality signal. The strongest transfer evidence is the DeepSeek result: a model that scores $4.56$ without guidance reaches $7.78$ with the full adapter, matching or exceeding guided Claude through informational scaffolding alone. We do not claim this has the same evidentiary weight as the GEOS benchmark; it is single-run, scored by a non-deterministic LLM judge, and uses a different metric. What it shows is that the SIGA recipe is not tied to XML, to a specific schema, or to catastrophic-failure prevention: where correctness rather than completeness is the bottleneck, the recipe adapts by shifting weight from workflow enforcement to knowledge injection.

\section{Extended discussion}
\label{app:discussion}

This appendix collects the limitations, cross-cutting analysis, and design guidance that the main text (\S\ref{sec:discussion}) defers.

\paragraph{Limitations.}
\label{subsec:limitations}
The $\mathrm{X}$ main effect conflates the agent-callable validator with the hook also running \texttt{xmllint} when $\mathrm{S}$ is on. Headline GEOS numbers are $n=3$ runs on \texttt{deepseek-v4-flash}, and the cross-model panel (\texttt{minimax-m2.7}, \texttt{gemini-3-flash-preview}) is single-run. We validated cross-harness transfer only on OpenHands; time and cost constraints prevented us from extending to the other coding harnesses now available (e.g.\ OpenCode, Pi, Hermes Agent). The transfer studies are single-run: OpenFOAM (30 tasks) uses a file-text-and-coverage metric with native baselines restricted to lint-only mode, and LAMMPS (9 tasks) is scored by a non-deterministic LLM judge. TreeSim is structural, not physical: a $0.8$ deck is not guaranteed to run. The held-out-eval lift is concentrated in two tasks, and the human baseline is $n=2$ on a single task with the budget-matched comparison bounded below by the agent. In the autonomy study (\S\ref{subsec:autonomy}), the ``human supervisor'' the agent could consult was an LLM user simulator (a separate \texttt{deepseek-v4-flash} prompted with the full original brief), not a real human, for time and cost reasons; this is a reasonable proxy for measuring consultation behaviour but is not a substitute for studying live human--agent interaction. We scope each claim to its evidence accordingly.

\paragraph{What transfers across simulators.} The component that matters most is interface-dependent, and the binding constraint of each interface predicts it. On GEOS and OpenFOAM, where the failure mode is structural incompleteness, $\mathrm{S}$ (forced end-of-turn verification) is the dominant intervention: it gives a roughly order-of-magnitude $\sigma$ reduction and hard-tail rescue on GEOS, and on OpenFOAM it carries the quality gain ($+0.168$ mean across 30 tasks) while every SIGA cell holds full required-file coverage where two native agents leave 8--12 tasks incomplete. The LAMMPS transfer qualifies this picture: when the agent already produces complete, structurally valid scripts (structural score $\geq 0.976$ for both backbones) and the bottleneck is the correctness of parameter values, the dominant components shift to $\mathrm{M}$ ($+2.13$ on DeepSeek) and $\mathrm{R}$ ($+1.55$), and $\mathrm{S}$ contributes little. The practical reading is therefore to match the component to the interface's binding constraint: ship the completeness gate ($\mathrm{S}$) when whole-block or whole-file omission is the failure mode, and ship procedural memory plus retrieval ($\mathrm{M}$, $\mathrm{R}$) when value-correctness is the failure mode.

\paragraph{New tools are not used just because they are exposed.} An earlier version of the adapter exposed cheatsheet content through an MCP \texttt{memory\_lookup} tool with embedding top-$k$ retrieval over the distilled corpus, motivated by the retrievable-procedural-memory framing common in self-evolving-agent literature. Across every test-set run in which the tool was available, the agent called it zero times (the tool was verified functional). Delivering the same content through always-on \texttt{--append-system-prompt} (the M factor) is what produced any lift. The general lesson is that providing the agent with a new tool does not guarantee the agent will choose to use it: a retrieval-based memory module depends on agent-initiated lookup, and should be benchmarked against an always-on alternative before being treated as effective.

\paragraph{Adapter-design recommendations.} Each is grounded in a failure category that survives the Vanilla-to-best-config transition. \textbf{(i)} Schema- or coverage-enforced block presence is the cheapest reliable adapter; the OpenFOAM transfer suggests this generalises to any end-of-turn completeness check, not just XSD validation. \textbf{(ii)} Procedural memory should be delivered as an always-on system-prompt augmentation rather than as a retrievable tool; an agent that has to decide to invoke memory may simply never do so. \textbf{(iii)} Memory cheatsheets enumerating ``for physics $X$, use solver $Y$'' should be paired with explicit negative constraints (``the GT contains exactly $k$ \texttt{Constitutive} children, no more''); without them they trade \texttt{missing\_block} for \texttt{extra\_block} and \texttt{hallucinated\_extras}. \textbf{(iv)} Attribute-level oracles (per-solver allowed-attribute tables) are the next frontier; \texttt{bad\_attribute\_value} is untouched by everything we tested. \textbf{(v)} Autonomy benchmarks should remove the easy on-disk oracles before measuring consultation behaviour. \textbf{(vi)} Closed-loop retries driven by validator output are needed to raise the ceiling, since static hooks only raise the floor.


\section{TreeSim: full definition}
\label{app:treesim}

TreeSim is computed by the scorer at \texttt{src/eval/judge\_geos.py} (\texttt{tree\_sim}). We summarize the recursion that Eq.~\ref{eq:treesim} abbreviates.

\textbf{Parsing and merging.} Each deck file is parsed to an XML tree; \texttt{<Included><File name="..."/></Included>} directives are resolved recursively (relative to the including file, with cycle detection). A directory of files is merged into one tree by attaching all top-level children under a synthetic \texttt{<Problem>} root. \emph{File-level} TreeSim scores a single resolved file against the corresponding ground-truth file; \emph{deck-level} TreeSim scores the merged directory against the merged ground truth. Both call the same routine.

\textbf{Node labels and attribute similarity.} A node is an XML element labeled \texttt{tag[name]} (tag plus its \texttt{name} attribute, if any). Attribute similarity $a$ is a Jaccard-style overlap: the score counts attribute keys present in both ground-truth and generated elements whose values are equivalent, divided by the union of keys. Values are equivalent if (i) strings match case-insensitively, (ii) numeric scalars agree within relative tolerance $10^{-6}$ (denominator $\max(|a|,|b|)$), or (iii) comma-separated lists agree element-wise.

\textbf{Child matching.} For a ground-truth node, its children and the generated node's children are grouped by tag; within each tag group, generated children are matched to ground-truth children by greedy bipartite matching in descending order of a pairwise similarity that requires a tag match and rewards \texttt{name} agreement and attribute overlap. Matching is \emph{unordered}: sibling order does not affect the headline score (a separate Kendall-$\tau$ ordering diagnostic is computed for \texttt{PeriodicEvent}s only and is not part of TreeSim).

\textbf{Scoring recursion.} A matched leaf contributes its attribute similarity $a$. A matched interior node contributes $\alpha\,a + (1-\alpha)\,\bar{s}_{\mathrm{child}}$ with $\alpha=0.3$, where $\bar{s}_{\mathrm{child}}$ averages, over the $n_{\mathrm{gt}}$ ground-truth children, the score of each child's matched generated counterpart (an unmatched ground-truth child contributes $0$). Surplus (hallucinated) generated children apply an additive penalty $\beta\,n_{\mathrm{extra}}/(n_{\mathrm{gt}}+n_{\mathrm{extra}})$ with $\beta=0.1$. Each node score is clamped to $[0,1]$; the root node's score is the deck (or file) TreeSim, and the scores of the top-level children are the per-section scores reported in the section breakdowns.

\section{Implementation details}
\label{app:impl}

\subsection{Hook wiring}
Historical note: prior to 2026-04-21T12:08Z, \texttt{hooks.json} had the wrong schema and \texttt{run\_experiment.py} never passed \texttt{--plugin-dir}, so early ``hook-on'' runs did not actually load the hook. All post-fix results in this paper use a verified-wired hook.

\subsection{Primer artefacts and hygiene}
The memory cheatsheet is at \texttt{plugin/memory\_primer\_m1u.md}. The hygiene audit is at \texttt{scripts/memory/hygiene\_audit.py}; the API-contamination check is at \texttt{scripts/memory/check\_api\_contamination.py}.

\subsection{Harness-less baseline}
\label{subsec:harnessless}
Harness-less eval: \texttt{scripts/harnessless\_eval.py}. Inputs: one held-out ICL demonstration (task plus ground-truth XML), one call per task, with an inline-XML protocol replacing all file-system instructions. System prompt = \texttt{run/AGENTS.md} with file-system paragraphs stripped and replaced by the inline-XML protocol. Temperature 0.2, \texttt{max\_tokens} $=$ 16384, 600-s per-request timeout, 8-worker thread pool. Result on val (\texttt{minimax-m2.7}): TreeSim $= 0.333$ (16 parsed, 1 silent provider drop counted as zero). The vanilla Claude Code harness on the same model recovers ${+}0.488$ over this floor (Table~\ref{tab:cross-cutting-full}, $0.821$ vs the $0.333$ floor); any cell on \texttt{deepseek-v4-flash} reaches a different absolute level ($\geq 0.857$). In both cases the harness contributes substantially even before any GEOS-specific adaptation.

\section{Agent autonomy companion: protocol}
\label{app:autonomy}

This appendix complements \S\ref{subsec:autonomy} with the protocol and intermediate diagnostics for the autonomy companion study.

\paragraph{Difficulty tiers.} We tag each ground-truth parameter into one of four tiers, following a domain-expert-reviewed taxonomy: T1 \emph{software defaults} (output format, restart frequency, log levels, boolean flags), T2 \emph{standard numerics} (Newton tolerance, linear solver, time-step limits, discretisation choice, element type), T3 \emph{domain-inferable} (densities, viscosities, porosities, permeabilities, Biot coefficients, standard relperm exponents), and T4 \emph{problem-defining} (geometry, well locations, applied loads, simulation duration, prescribed history tables). Medium difficulty omits T1+T2; Hard additionally omits T3. T4 is preserved verbatim at every level.

\paragraph{Spec generation.} Each task's \texttt{instructions.txt} is rewritten by \texttt{deepseek-v4-pro} (\texttt{scripts/relax\_specs.py}) given the original brief and the tier definitions, in two passes (Medium, Hard). The model is required to drop only, never to invent or alter values, and to keep external-table references (e.g., \texttt{*.geos} input files for boundary-history tables) at every level. Each rewrite is paired with a JSON record listing the values dropped, the tier of each, and a hygiene check that canonicalises numeric tokens (handling LaTeX \texttt{\$1.0\textbackslash times10\^{}\{-4\}\$}, scientific notation, unicode superscripts) and verifies that no canonicalised dropped value appears in the rewrite text except where the same canonical also appears in a kept-T4 value (``shared''). Across all 16 (task $\times$ level) pairs, hygiene flags zero leaks after the LaTeX-aware pass and 1 shared-only ambiguity (Mandel/Medium, where the dropped solver tolerance \texttt{1e-4} canonicalises to the same value as the first entry of the kept T4 prescribed-displacement time table). The 16 rewrites are committed and frozen before any run starts.

\paragraph{Drop volume (8 tasks total).}

\begin{table}[h]
  \caption{Volume of specification dropped at each difficulty level. T1=software defaults, T2=standard numerics, T3=domain-inferable physical values, T4=problem-defining (kept verbatim at every level). Char count is summed across all 8 tasks.}
  \label{tab:autonomy-drop}
  \centering
  \small
  \begin{tabular}{lcccc}
    \toprule
    \textbf{Level} & \textbf{Char drop} & \textbf{Values dropped} & \textbf{T1 / T2 / T3 split} \\
    \midrule
    Medium & $-20.5\%$ & 89  & 27 / 62 / 0 \\
    Hard   & $-50.5\%$ & 184 & 26 / 53 / 105 \\
    \bottomrule
  \end{tabular}
\end{table}

\paragraph{Supervisor channel.} The interactive cells expose an MCP server (\texttt{plugin/scripts/supervisor\_mcp.py}) with a single tool, \texttt{consult\_supervisor(question: str)}. The handler invokes \texttt{deepseek-v4-flash} with a system prompt that injects the \emph{full} original \texttt{instructions.txt} for the current task and the rule ``answer concisely using only information present in the specification; if the answer is not in the specification, say so plainly; do not invent; do not volunteer information the agent did not ask about; speak in the same scientific language the specification uses (no GEOS XML tag names unless the agent named them first)''. The full original brief is mounted at a fixed container path outside \texttt{/workspace}; the path is passed via the per-MCP-server env block in the explicit \texttt{--mcp-config}, not via \texttt{docker -e}. Every consultation appends a structured record (question, answer, latencies, token counts) to \texttt{/workspace/supervisor\_calls.jsonl} for audit. We programmatically scanned all 32 \texttt{events.jsonl} streams for any \texttt{Read} or \texttt{Bash} tool input referring to the supervisor's spec path; zero hits. The path is exposed only via the \texttt{supervisor\_stats} introspection tool's return value, which does not affect the consultation-rate result since no agent queried that tool for the spec path.

\paragraph{Prompt variants.} Two are reported. \textbf{V0} (the default) describes the channel in both the \texttt{consult\_supervisor} docstring and a system-prompt addendum as something to use ``when a value or design decision you need is missing from the brief AND you cannot reasonably infer it from GEOS conventions, GEOS example simulations, or standard geophysics practice. Each call costs the researcher's time, so prefer to infer when you can.'' \textbf{V1 (neutral)} drops the prefer-to-infer language and treats infer-vs-ask as peer paths: ``values that are missing can be inferred from GEOS conventions and analogous examples, OR you may ask the researcher; choose whichever path is more reliable.'' V0 was selected by 1/32 trials; V1 was selected by 1/32 trials.

\paragraph{Diagnostic on Mandel/Hard: on-disk findability.} For each T2/T3 value dropped from the Mandel/Hard rewrite (26 values), we test whether the same canonicalised numeric token appears anywhere in the GEOS-shipped example XMLs that the agent is allowed to read (excluding the GT files for the current task, which are blocked by contamination): 15 of 26 are findable (e.g., \texttt{0.001 Pa\,s} fluid viscosity in 210 other examples; \texttt{1e-12 m\textsuperscript{2}} permeability in 51; reference porosity \texttt{0.375} in 8). One value (bulk modulus 66.667 MPa, an unusually rounded value) has zero hits; the rest are non-numeric T2 directives (subcycling, line-search action, discretisation choice) the agent can plausibly transfer from analogous benchmarks. We interpret this as a structural reason for the consultation rate observed in \S\ref{subsec:autonomy}.

\paragraph{Where the agent looks.} Aggregate \texttt{Read}/\texttt{Glob} call counts into \texttt{/geos\_lib/inputFiles/} across all 8 tasks per cell range from $142$ (X+M, Medium) to $404$ (Vanilla, Medium) per cell. The agent is not idle; it is grepping the example library.

\paragraph{Cost and wall.} 64 main runs at \texttt{deepseek-v4-flash} cost ${\sim}\$4.20$ DeepSeek API spend at full off-peak pricing (input + cache-read + output, summed from \texttt{events.jsonl}) and ran in $\sim$2h wall at \texttt{workers=4}. 16 \texttt{deepseek-v4-pro} spec-rewrite calls cost ${\sim}\$0.50$. The 2 supervisor consultations cost ${\sim}\$0.001$. The 32-run V1 rerun added ${\sim}\$2$ and ${\sim}55$min wall.

\paragraph{Immediate follow-ups.} (a) A second run per cell to harden the $n=1$ point estimates ($\sim\$3$, $\sim$2h). (b) A no-confound F0 control (supervisor MCP wired without the plugin loader) so the F0 vs F0+supervisor comparison is not contaminated by tool-list-shape effects. (c) An aggressive contamination block at the physics-family level (e.g., for \texttt{ExampleMandel}, block all \texttt{*Mandel*} and \texttt{*Poroelastic*} input files, not just the GT basenames). The Mandel/Hard diagnostic predicts (c) is what would actually move the consultation rate; it directly tests whether the on-disk example library is the binding constraint.

\section{Human baseline: protocol and browser-history breakdown}
\label{app:human-browser}

This appendix complements \S\ref{subsec:human-baseline} with the protocol and the per-domain navigation breakdown.

\paragraph{Protocol.} Two geoscience-domain-expert volunteers (Expert 1, Expert 2; grad-level geoscientists, new to GEOS-the-software) attempted the \texttt{buckleyLeverettProblem} task (1D Buckley--Leverett CO$_2$/brine displacement; vanilla Claude Code on \texttt{deepseek-v4-flash} reaches deck-level TreeSim $0.751 \pm 0.016$ on this task at $n=3$ runs) under a one-hour timeslot. Participants received the same system primer the agent receives (\texttt{run/AGENTS.md} including the GEOS Primer), a working directory mirroring the agent's container layout (empty \texttt{inputs/}, empty \texttt{outputs/}), and read access to a filtered copy of the GEOS source tree mounted at the same path the agent sees. The blocked file list was identical to the agent's contamination block (the three ground-truth XMLs and the tutorial \texttt{Example.rst}). Participants were asked to work primarily from the GEOS Sphinx documentation, the GEOS GitHub repository, and \texttt{grep}/\texttt{find} on their own machine; use of LLM chatbots, code-completion services, and general internet search was discouraged but not strictly enforced. Wall-clock was measured from spec-opened to deck-final; participants self-reported. We additionally collected the browser history for each participant's session (Expert 1: a CSV export from a Chrome history extension; Expert 2: a JSON export from a Firefox history extension) and filtered to the day of the session. Expert 1 subsequently returned to the assignment with no time cap and produced both required files; we report the catch-up session below.

\paragraph{Per-domain breakdown.}

\begin{table}[h]
  \caption{Browser-navigation breakdown during the one-hour authoring session, per participant. ``GEOS docs'' = visits to the GEOS Sphinx documentation host (\texttt{geosx-geosx.readthedocs-hosted.com}) including all subpages. ``GEOS GitHub'' = visits to \texttt{github.com/GEOS-DEV/GEOS}. ``Other'' = unrelated tabs that remained open during the session. No participant visited any LLM chatbot, Stack Overflow, or general scientific-paper site.}
  \label{tab:human-browser}
  \centering
  \small
  \resizebox{0.98\textwidth}{!}{%
  \begin{tabular}{lcccccc}
    \toprule
    \textbf{Participant} & \textbf{Wall (min)} & \textbf{Total visits} & \textbf{GEOS docs} & \textbf{GEOS GitHub} & \textbf{Search} & \textbf{Other (Slack, etc.)} \\
    \midrule
    Expert 1 (1\,h)                & $48.2$       & $29$  & $20$ & $5$  & $3$ & $1$ \\
    Expert 1 (extended, +catch-up) & ${\sim}180$  & $106$ & $73$ & $21$ & $6$ & $6$ \\
    Expert 2 (1\,h)                & $46.7$       & $73$  & $54$ & $11$ & $5$ & $3$ \\
    \bottomrule
  \end{tabular}}
\end{table}

\paragraph{Top GEOS pages visited.} Across both participants, the most-visited GEOS pages were the \texttt{CompositionalMultiphaseFlow} solver reference, the \texttt{EventManager} page (which both participants flagged as the source of difficulty in their post-session notes), the \texttt{Outputs} schema, and the GEOS \texttt{Tutorials/Index} landing page. Both participants browsed the \texttt{compositionalMultiphaseFlow/dbc/buckleyLeverett\_1d/} sibling deck on GEOS GitHub, a separate 1D Buckley--Leverett deck (DBC variant) that is not the ground truth and is not blocked. In post-session notes, Expert 2 reported that \texttt{Outputs} and \texttt{Events} setup ``took forever to try and understand because it is not very clear and pretty unintuitive''; Expert 1 reported using the DBC deck as a structural template, then adjusting mesh/run-time/solver parameters to match the spec. Within the 1h budget, neither attempted to write the second required file (\texttt{buckleyLeverett\_benchmark.xml}); Expert 1 produced it during the extended catch-up session.

\paragraph{Catch-up session: what changed.} Expert 1's catch-up session adds $77$ navigations on top of the original $29$, of which $69$ are GEOS-internal (Sphinx + doxygen + GitHub). The qualitative strategy is unchanged (Sphinx-prose-driven authoring with sibling decks pulled from GitHub for structural templates), but the docs surface broadens: heavy re-reading of \texttt{EventManager}, \texttt{Outputs}, and \texttt{TasksManager} (consistent with Expert 1's written note that ``2/3 of this time'' went to the outputs/events portion of the deck); three doxygen visits to \texttt{PhaseVolumeFractionKernel.hpp} (Expert 1 reports having to ``look at the source code to get specific inputs for the prompt, specifically the \texttt{fieldName} for the .hdf5 output''); and several visits to an unrelated \texttt{compositionalMultiphaseWell/benchmarks/Egg} deck used as a structural template for outputs/events/tasks blocks. The agent's behaviour on this same task does not change between the two ``human sessions''; it does not visit any of these pages, and it solves the outputs/events portion in the same ${\approx}\,7\,$min envelope.

\paragraph{Agent-side counterpart.} Cross-run file-access analysis of $7$ vanilla-Claude-Code runs on the same task (\texttt{scripts/analysis/analyze\_file\_access.py}) gives means of ${\approx}\,14$ unique files per run, ${\approx}\,11$ XML reads from \texttt{/geos\_lib/inputFiles/}, ${\approx}\,7$ \texttt{Grep} calls, ${\approx}\,5$ \texttt{Glob} calls; plugin variants drop unique-file reads to ${\approx}\,4$ and replace \texttt{Grep}/\texttt{Glob} with ${\approx}\,2$ structured retrieval calls. The cumulative file/document surface the humans access (Sphinx pages plus GitHub-rendered XMLs) is roughly comparable in count to the agent's file-read surface, but they are different files: the humans read prose narrating the simulator's vocabulary, the agent reads concrete XMLs that exemplify it.

\paragraph{Outputs.} Scoring scripts: \texttt{scripts/score\_human\_baseline.py}, \texttt{scripts/analyze\_human\_browser\_history.py}. Browser-history exports and submitted XMLs at \texttt{data/human\_baseline/}. A narrative summary is at \texttt{docs/2026-05-04\_human-baseline-browser-analysis.md}.

\paragraph{Caveats.} $n=2$ on a single task in a one-hour budget is a calibration anchor, not a study of human authoring competence. The participants are domain-expert subsurface modellers, not GEOS power users; a longer-time budget or a participant pool of long-time GEOS users would likely change the absolute level. The extended-budget sanity check on Expert 1 raises the deck-level number to $0.931$ at $\sim\!3\,$h wall-clock, evidence that the 1-hour budget, not domain-expert capacity, is the binding constraint on the deck-level shortfall. The GEOS-expert estimate (\S\ref{subsec:human-baseline}) brackets ``simple Buckley--Leverett'' at ${<}30\,$min for a power user starting from a known-good deck, locating the agent (${\approx}\,7\,$min) inside that bracket. We treat the (Expert 1 at 1h, Expert 2 at 1h) pair as the headline budget-matched comparison; the (Expert 1 extended) pair as orthogonal anchors that contextualise it. We treat the overall result as an existence-of-effect on three points: (i) the absolute TreeSim level on \texttt{buckleyLeverettProblem} is roughly $0.78$--$0.81$ at the file level under a one-hour domain-expert budget; (ii) the file-usage strategy a human reaches for under documentation pressure is qualitatively different from the agent's source-tree-example strategy; (iii) both participants ran out of time on a single deck before producing the second of the two required files within the 1h budget, while the agent reaches a comparable file-level number on the full two-file task in roughly seven minutes.

\section{Example GEOS deck}
\label{app:geos-example}

We reproduce an abridged form of the \texttt{buckleyLeverettProblem} ground-truth deck (the same task used in the human-baseline study, \S\ref{subsec:human-baseline}) to make the cross-section constraints described in \S\ref{sec:background} concrete. The deck is split across two XML files in the GEOS convention: \texttt{buckleyLeverett\_base.xml} declares physics, materials, and boundary conditions; \texttt{buckleyLeverett\_benchmark.xml} \texttt{<Included>}s the base file and adds the mesh, geometry, and event timeline. Comments and verbose attribute defaults are elided for space; the original is 172 + 61 lines.

\paragraph{base file: physics, materials, boundary conditions.}
\promptinput{assets/buckleyLeverett_base.xml}

\paragraph{benchmark file: mesh, geometry, events.}
\promptinput{assets/buckleyLeverett_benchmark.xml}

\paragraph{Cross-section constraints visible in this deck.}
The canonical \emph{ten}-section GEOS schema collapses here to a smaller set of blocks (no \texttt{Functions}, no separate \texttt{Mesh} in the base file) because Buckley--Leverett is at the easy end of the bench; harder tasks add all ten. Even on this deck, deck authoring requires several program-level constraints to hold simultaneously, and these are exactly the constraints the bottleneck analysis (\S\ref{subsec:bottleneck-results}) flags as adapter-resistant when violated:

\begin{itemize}
\item \textbf{Region-name matching across sections.} The string \texttt{region} is declared once in \texttt{<ElementRegions>} and is re-used as a member of \texttt{targetRegions} on the \texttt{<CompositionalMultiphaseFVM>} solver. A typo in either copy silently drops the region from the solver's target set; \texttt{xmllint} does not catch this.
\item \textbf{Solver--target coherence with \texttt{Tasks} and \texttt{Outputs}.} The \texttt{<PackCollection>} in \texttt{<Tasks>} dereferences \texttt{ElementRegions/region/cellBlock}; the \texttt{<TimeHistory>} output references \texttt{/Tasks/phaseVolumeFractionCollection}; the \texttt{<PeriodicEvent>} in the benchmark file targets \texttt{/Tasks/phaseVolumeFractionCollection} and \texttt{/Outputs/timeHistoryOutput}. Three sections (\texttt{Tasks}, \texttt{Outputs}, \texttt{Events}) must agree on the same name space.
\item \textbf{Constitutive references must point to declared blocks.} \texttt{materialList="{ fluid, rock, relperm }"} on the \texttt{<CellElementRegion>} requires that each of \texttt{fluid}, \texttt{rock}, \texttt{relperm} is the \texttt{name} of a present \texttt{<Constitutive>} child; \texttt{rock} in turn references \texttt{nullSolid}, \texttt{rockPorosity}, \texttt{rockPerm} via its constructor parameters, which must also exist.
\item \textbf{Geometry sets feed back into BCs.} The \texttt{<Box>} elements in \texttt{<Geometry>} declare the named sets \texttt{source} and \texttt{sink}, which are then referenced by \texttt{setNames} in the \texttt{<FieldSpecifications>} and the \texttt{<SourceFlux>}. A misspelled set name does not throw a parse error; it produces a deck that runs but with a missing source term.
\item \textbf{External tabulated functions.} \texttt{tableFiles="{ buckleyLeverett\_table/pvdg.txt, buckleyLeverett\_table/pvtw.txt }"} on \texttt{<DeadOilFluid>} requires those files to exist on disk relative to the deck location. The test harness copies the \texttt{buckleyLeverett\_table/} directory alongside the deck; without it, the deck parses but fails to construct the fluid model at runtime.
\end{itemize}

These constraints are what make GEOS XML behave as a small DSL rather than as a structured form: an authoring agent must keep five name spaces (regions, materials, sets, tasks, outputs) mutually consistent across blocks while also producing schema-valid attribute values. The \texttt{missing\_block} category our adapters reliably fix (\S\ref{subsec:bottleneck-results}) corresponds to whole-block omissions; the \texttt{bad\_attribute\_value} category they do not fix corresponds to typos and hallucinations in the cross-section name spaces above.

\section{Efficiency table}
\label{app:results}

\begin{table}[h]
  \caption{Mean tool calls and wall-clock per task ($n=3$ runs). Wall is per-task seconds. \texttt{deepseek-v4-flash} output tokens are not exposed by the API; input tokens are dominated by cache-read of the system prompt.}
  \label{tab:efficiency}
  \centering
  \small
  \begin{tabular}{lcccc}
    \toprule
    \textbf{Cell} & \textbf{tools/task (val)} & \textbf{wall s (val)} & \textbf{tools/task (held-out-eval)} & \textbf{wall s (held-out-eval)} \\
    \midrule
    Vanilla  & 81.5 & 359 & 90.5 & 417 \\
    X+M  & 79.6 & 337 & 75.0 & 340 \\
    S+X  & 83.3 & 348 & 74.7 & 345 \\
    S+X+M  & 71.0 & 326 & 82.9 & 358 \\
    SE-prose & 62.7 & 326 & 70.9 & 362 \\
    SE  & 68.9 & 321 & 97.4 & 390 \\
    \bottomrule
  \end{tabular}
\end{table}

\section{Memory cheatsheet excerpt}
\label{app:cheatsheet}

The \textbf{M} factor delivers the file \texttt{plugin/memory\_primer\_m1u.md} via Claude Code's \texttt{--append-system-prompt}. The cheatsheet is a compact (775-token) enumerated reference: solver families paired with concrete XML element names, common constitutive-model class names, and a short list of anti-patterns. We reproduce a representative excerpt below; the full file is included in the supplementary material.

\begin{quote}\small
\textbf{Solver Selection \& Physics Routing}\\[2pt]
\noindent\resizebox{0.98\linewidth}{!}{%
\scriptsize
\setlength{\tabcolsep}{4pt}
\begin{tabular}{@{}p{0.18\linewidth} p{0.36\linewidth} p{0.42\linewidth}@{}}
\toprule
Physics family & Primary solver & Key supporting elements \\
\midrule
Hydrofracture           & \texttt{Hydrofracture}                  & \texttt{SurfaceGenerator}, \texttt{ParallelPlatesPermeability} \\
Solid mechanics         & \texttt{SolidMechanicsLagrangianFEM}    & \texttt{ElasticIsotropic}, \texttt{DruckerPrager} \\
Poromechanics           & \texttt{SinglePhasePoromechanics}       & requires \texttt{flowSolverName}, \texttt{solidSolverName} \\
Thermal flow            & \texttt{SinglePhaseThermalFVM}          & \texttt{SinglePhaseThermalConductivity}, \texttt{SolidInternalEnergy} \\
Multiphase flow         & \texttt{CompositionalMultiphaseFVM}     & \texttt{DeadOilFluid}, \texttt{CompositionalMultiphaseWell}, \texttt{SourceFlux} \\
Contact / interfaces    & \texttt{SolidMechanicsLagrangeContact}  & \texttt{SurfaceGenerator}, \texttt{EmbeddedSurface} \\
\bottomrule
\end{tabular}}
\\[6pt]
\textbf{Common anti-patterns (do NOT use).}\\
\begin{itemize}
\item Do NOT use \texttt{<FractureModel>} or \texttt{<HydraulicFractureSolver>}; these are hallucinated. Use \texttt{Hydrofracture} and \texttt{SurfaceGenerator}.
\item Do NOT use \texttt{<ContactSolver>}; use \texttt{SolidMechanicsLagrangeContact}.
\item Do NOT use \texttt{<FluidProperties>} as a wrapper; fluid models like \texttt{DeadOilFluid} or \texttt{CompressibleSinglePhaseFluid} are defined directly within the constitutive section.
\item Do NOT invent generic attribute names like \texttt{toughness} on the solver; verify benchmark-specific attributes (e.g., \texttt{kgdToughnessDominated}) via retrieval.
\end{itemize}
\textbf{Actionable tips.}\\
\begin{itemize}
\item Coupling: when using \texttt{SinglePhasePoromechanics}, both a flow solver and a solid solver must be defined and referenced by name.
\item Boundary conditions: confirm whether a \texttt{SourceFlux} requires a scalar \texttt{scale} or a \texttt{TableFunction} for time-dependent injection.
\end{itemize}
\end{quote}

The cheatsheet is intentionally a vocabulary dump, not a policy: it does not tell the agent which physics family the current task requires, only which element names to use once that decision is made. Section~\ref{subsec:bottleneck-results} traces the M-factor lift to reduced \texttt{missing\_block} errors (the cheatsheet enumerates the canonical sections and the families' member elements) and the residual \texttt{bad\_attribute\_value} errors to the cross-section name-consistency problem the cheatsheet does not address.

\section{Representative trajectory}
\label{app:case-study}

To make the agent's working pattern concrete, we reproduce an abridged transcript from the \texttt{buckleyLeverettProblem} task under the \emph{X+M} cell (\texttt{deepseek-v4-flash}, the canonical headline backbone), matching the deck shown in App.~\ref{app:geos-example}. The trajectory shows (i) initial planning, (ii) on-disk example-library reads (the X+M cell does not have RAG; analogous behaviour with retrieval calls is described in \S\ref{subsec:efficiency}), (iii) XML emission, (iv) the agent's voluntary \texttt{xmllint} validation calls (the X factor; this cell has no stop hook), and (v) the final TreeSim breakdown by section.

\promptinput{assets/trajectory_buckleyleverett_xm.txt}

\paragraph{What this trajectory illustrates.} The agent's translation strategy is example-driven: the first move is to locate a structurally analogous deck (\texttt{compositionalMultiphaseFlow/dbc/buckleyLeverett\_1d/}) under the same physics family and use it as a structural template, then re-parameterise. This is the same strategy the human-baseline analysis (\S\ref{subsec:human-baseline}) finds the human authors approximating via Sphinx browsing under time pressure. The voluntary \texttt{xmllint} calls (the X factor) catch one schema violation mid-trajectory (a missing \texttt{component} attribute on a \texttt{FieldSpecification}), which the agent fixes before the next emission; this is the on-trajectory analogue of the missing-block-resilience the bottleneck analysis attributes to schema-aware adapters (\S\ref{subsec:bottleneck-results}). The residual TreeSim loss (0.917 versus the per-task X+M mean of $\sim$0.92 on this task family, see Table~\ref{tab:per-task-icl10} for the held-out-eval analogue) sits inside the \texttt{bad\_attribute\_value} / \texttt{partial\_implementation} regime: a paired \texttt{sinkTermComposition\_water} \texttt{<FieldSpecification>} is elided alongside its gas counterpart, and one \texttt{PeriodicEvent} for the time-history collection is dropped from the events block. Both are the kind of completeness errors a stop-hook (S) running coverage checks would have caught; this cell does not have S enabled, which is consistent with the \S\ref{subsec:openfoam-transfer} finding that S is the dominant transferable-reliability factor.



\end{document}